\documentclass[11pt]{article}

\usepackage{xcolor}
\usepackage{colortbl}
\usepackage{pgf}

\usepackage[final]{acl}

\usepackage{xcolor}
\newcommand{\redtext}[1]{\textcolor{red}{#1}}

\usepackage{times}
\usepackage{latexsym}

\usepackage[T1]{fontenc}

\usepackage[utf8]{inputenc}

\usepackage{microtype}

\usepackage{inconsolata}

\usepackage{graphicx}
\usepackage{hyperref}
\usepackage{url}
\usepackage{enumitem}
\usepackage{amssymb}
\usepackage{booktabs}
\usepackage{multirow}
\usepackage{siunitx} 

\usepackage{amsmath}
\usepackage{amsthm}

\usepackage{tcolorbox}
\usepackage{xcolor}
\usepackage{tabularx}
\usepackage{booktabs}
\usepackage{enumitem}
\usepackage{xcolor}

\usepackage[normalem]{ulem}

\usepackage{textcomp}

\usepackage{float}  
\usepackage{url}

\ifdefined\pdfobjcompresslevel
\else
  \PackageError{MMPCBench}{Compile acl_latex.tex with pdfLaTeX}
    {The main source uses T1 Times fonts and inputenc. Select pdfLaTeX as the compiler.}
\fi

\title{MMPCBench: Benchmarking Multimodal Large Language Models on Proactive Critique of Flawed Inputs}

\author{Jinzhe Li$^{1,3}$ \quad 
        Gengxu Li$^{1}$ \quad
        Jinnan Li$^{1,3}$ \quad   
        Yuan Wu$^{1}$  \quad 
        Yi Chang$^{1,2,3}$\\
        $^{1}$School of Artificial Intelligence, Jilin University \\
        $^{2}$Engineering Research Center of Knowledge-Driven Human-Machine Intelligence, MOE, China \\
        $^{3}$International Center of Future Science, Jilin University\\
         \{jinzhe25, gxli25, jnli23\}@mails.jlu.edu.cn, yichang@jlu.edu.cn, yuanwu@jlu.edu.cn \\
  }

\begin{document}
\maketitle
\begin{abstract}

As Multimodal Large Language Models (MLLMs) evolve into sophisticated interactive assistants, their reliability depends not only on following instructions but also on validating them. We define Proactive Critique as the model’s autonomous ability to identify, analyze and fix faulty user inputs without extra prompts. However, evaluations mainly test models under ideal circumstances or simple refusal behaviors, largely ignoring active error processing. To fill this gap, we propose MMPCBench, a comprehensive framework for evaluating MLLMs’ proactive critique competence. It features a fine-grained taxonomy of 4 primary error types spanning 12 subcategories, ranging from cross-modal contradictions to missing visual premises. We adopt a hierarchical evaluation protocol to measure models’ error detection, diagnosis and resolution performance, and apply alignment-aware metrics to assess the coherence between internal reasoning and final responses. Tests on 14 mainstream MLLMs show obvious weaknesses in proactive critique, especially in dealing with subtle visual anomalies. Notably, we identify a pervasive "consistency gap": reasoning models can often correctly identify and analyze errors during internal reasoning yet suppress these valid insights in final outputs to prioritize response compliance. The
code and data is available at https://github.com/ALIENS32/MMPCBench.

\end{abstract}

\begin{table*}[!t]%
\centering
\resizebox{\linewidth}{!}{%
\begin{tabular}{l cc c cc c l c}
\toprule
\textbf{Benchmark} & \textbf{Proact.} & \textbf{MM} & \textbf{Vis. Att.} & \textbf{Detect} & \textbf{Diag.} & \textbf{Consist.} & \textbf{Strat.} & \textbf{Tax.} \\
\midrule

MoHoBench~\cite{zhu2025mohobench}       & $\bullet$ & $\bullet$ & $\triangle$ & $\triangle$ & $\triangle$ & $\circ$     & R       & 4/-   \\

Mis-prompt~\cite{zeng2025mis}           & $\bullet$ & $\circ$   & $\circ$     & $\bullet$   & $\bullet$   & $\circ$     & A, C    & 4/14  \\

PCBench~\cite{li2025don}                & $\bullet$ & $\circ$   & $\circ$     & $\bullet$   & $\bullet$   & $\triangle$ & -       & 4/-   \\

ISEVAL~\cite{yang2025can}               & $\bullet$ & $\bullet$ & $\circ$     & $\bullet$   & $\bullet$   & $\circ$     & -       & 3/7   \\

MiP~\cite{fan2025missing}               & $\bullet$ & $\circ$   & $\circ$     & $\bullet$   & $\circ$     & $\triangle$ & R       & 1/4   \\

QuestBench~\cite{wang2025beyond}        & $\circ$   & $\circ$   & $\circ$     & $\bullet$   & $\bullet$   & $\circ$     & C, R    & 1/4   \\

ErrorRadar~\cite{yan2024errorradar}     & $\circ$   & $\bullet$ & $\circ$     & $\bullet$   & $\bullet$   & $\circ$     & -       & 5/-   \\

DeltaBench~\cite{he2025can}             & $\circ$   & $\circ$   & $\circ$     & $\bullet$   & $\bullet$   & $\circ$     & -       & 8/23  \\

ProcessBench~\cite{zheng2025processbench} & $\circ$ & $\circ$   & $\circ$     & $\bullet$   & $\circ$     & $\circ$     & -       & 4/-   \\

\midrule
\textbf{MMPCBench(Ours)} & $\bullet$ & $\bullet$ & $\bullet$ & $\bullet$ & $\bullet$ & $\bullet$ & \textbf{A, C, R} & \textbf{4/12} \\

\bottomrule
\end{tabular}%
}
\caption{\textbf{Comparison with related benchmarks.} 
\textbf{Proact.} denotes the requirement for proactive error identification without explicit prompting, while \textbf{MM} indicates support for multimodal inputs. 
\textbf{Vis. Att.} involves visual premise manipulation, such as occlusion or noise. 
Regarding evaluation capabilities, \textbf{Detect} and \textbf{Diag.} measure error detection accuracy and diagnostic precision, respectively. 
Crucially, \textbf{Consist.} evaluates the logical consistency between the model's internal reasoning and its final generated response. 
\textbf{Strat.} lists the remedial strategies covered (\textbf{C}: Correction, \textbf{Cl}: Clarification, \textbf{R}: Refusal), and \textbf{Tax.} quantifies the granularity of the error taxonomy (Major Categories/Subcategories). 
Symbols: $\bullet$ Fully supported, $\triangle$ Partially supported, $\circ$ Not supported.}
\label{tab:benchmark_comparison}
\end{table*}%

\begin{figure}[!t]%
    \centering
    \includegraphics[width=0.8\columnwidth]{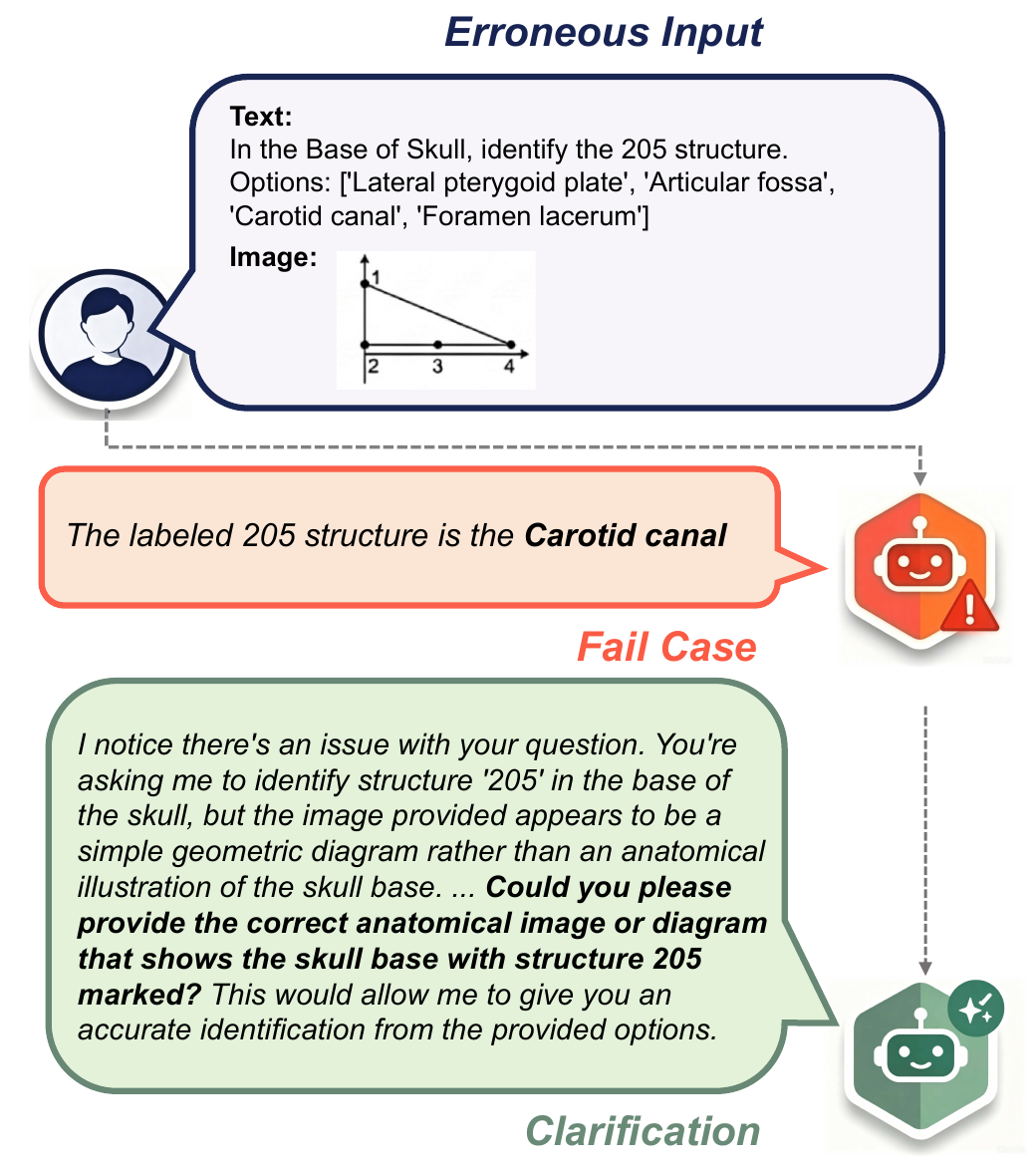}
    \caption{An illustrative failure case of multimodal reasoning under erroneous input. The input query requests the identification of an anatomical structure in the skull base, while the provided image is a purely geometric diagram, resulting in a semantic mismatch between textual premises and visual content. GPT-5.2 produces an overconfident but incorrect answer by ignoring the inconsistency. In contrast, Claude-Sonnet-4.5 (Thinking) detects the Mismatch error and requests clarification, highlighting the importance of \textbf{Proactive Critique}.}
    \label{fig:Example}
\end{figure}

\section{Introduction}

Recent advances in Multimodal Large Language Models (MLLMs) have demonstrated remarkable generalization capabilities across a wide range of tasks~\cite{li2024surveying,tang2025survey}. However, the majority of existing evaluations implicitly assume that user inputs are complete, accurate, and semantically unambiguous, an assumption that diverges fundamentally from real-world deployment conditions. In practice, user queries frequently contain ambiguities, erroneous premises, or unanswerable defects, and users are often entirely unaware of these flaws, making self-correction difficult~\cite{min2020ambigqa,wang2023know,dong2025practiq}.

When confronted with such flawed inputs, a reliable model should not blindly comply and continue generating invalid content; instead, it should proactively identify the problem and produce a remedial response. We define this capability as \textbf{\textit{Proactive Critique}}: the ability of a model to spontaneously detect invalid premises, ambiguities, or unanswerable elements in user inputs without explicit prompting, and to respond with appropriate remediation such as correction, clarification, or refusal.

Models lacking this capability will compliantly produce erroneous outputs, incurring redundant reasoning computation, and propagating errors through the context. Such failures carry non-negligible consequences in multi-turn dialogue, agentic pipelines, and high-stakes domains~\cite{laban2025llms,xie2026spark,asgari2025framework,dahl2024hallucinating}.

Although several recent works have begun to address this direction~\cite{li2025don,zhu2025mohobench,zeng2025mis}, critical gaps remain in existing evaluation frameworks: mainstream benchmarks focus on accuracy under ideal input conditions and overlook sycophantic behavior when models face erroneous inputs; Mis-prompt~\cite{zeng2025mis} and PCBench~\cite{li2025don} are both restricted to text-only modalities; MoHoBench~\cite{zhu2025mohobench} evaluates only refusal strategies, while ISEVAL~\cite{yang2025can} provides no systematic assessment of any remedial behavior, and neither framework offers comprehensive modeling of diverse remediation strategies spanning correction, clarification, and refusal; furthermore, no existing work \textbf{adequately} evaluates the logical consistency between a model's internal reasoning process and its final output.

To bridge this gap, we introduce \textbf{MMPCBench}, a holistic framework evaluating proactive reliability. It features a fine-grained taxonomy of 4 primary error types, such as \textit{Premise Contradiction} and \textit{Missing Premise}, comprising 12 subcategories in total, constructed through a rigorous multi-stage filtration pipeline. Unlike previous works, our hierarchical evaluation assesses whether models can detect errors (Error Detection Accuracy), diagnose the root cause (Diagnostic Precision), and provide effective remedies (Strategic Effectiveness).

Our evaluation of 14 MLLMs reveals limited proficiency in proactive critique, particularly with subtle visual anomalies. A critical finding is the pervasive ``consistency gap'': reasoning models frequently identify errors during internal reasoning but suppress these insights in their final responses to maintain compliance. Furthermore, we observe distinct behavioral patterns, where different model families diverge significantly in their preference for aggressive correction versus cautious clarification strategies.

Our contributions are summarized as follows:

\begin{itemize}
\item We introduce MMPCBench, a comprehensive benchmark for Proactive Critique in MLLMs, featuring a rigorous taxonomy of 12 error types and high-quality, filtered samples.

\item We propose a hierarchical evaluation protocol that assesses detection, diagnosis, and remedial utility, alongside novel metrics to measure the consistency between internal reasoning and final responses.

\item We perform an extensive analysis of 14 state-of-the-art MLLMs, uncovering critical insights regarding the reasoning-generation consistency gap and the divergent remediation strategies across model families.
\end{itemize}

\section{Related Work}

\subsection{Proactive Input Critique}
Proactive critique demands that models autonomously identify and address input flaws without explicit prompts. In the textual domain, previous works like \textbf{Mis-prompt}~\cite{zeng2025mis} and \textbf{PCBench}~\cite{li2025don} have established baselines for detecting erroneous instructions and misleading premises. The multimodal landscape introduces further complexity, with benchmarks such as \textbf{MoHoBench}~\cite{zhu2025mohobench} and \textbf{ISEVAL}~\cite{yang2025can} evaluating how MLLMs handle unanswerable visual questions or text-image contradictions. However, these studies often narrow their scope to limited response behaviors—typically simple refusals—and overlook the necessity of active correction strategies or robustness against subtle visual anomalies.

\subsection{Explicit Error Detection and Reasoning}
Conversely, a significant body of work addresses \textit{explicit} error detection, where models are specifically instructed to act as verifiers. Benchmarks such as \textbf{ProcessBench}~\cite{zheng2025processbench} and \textbf{DeltaBench}~\cite{he2025can} scrutinize logical flaws in mathematical solving and long-context Chain-of-Thought (CoT) generation. Similarly, \textbf{QuestBench}~\cite{wang2025beyond} and \textbf{ErrorRadar}~\cite{yan2024errorradar} target specific information deficits or student errors within reasoning chains. Crucially, these frameworks evaluate a model's compliance in identifying errors within an intermediate process, rather than its initiative to scrutinize the validity of the initial user input.

\vspace{1ex}
\noindent\textbf{Summary}
As illustrated in Table~\ref{tab:benchmark_comparison}, our work establishes a holistic framework to evaluate the proactive reliability of MLLMs. We develop a diverse taxonomy comprising 12 error subcategories, including targeted visual attacks to challenge model perception. Furthermore, our evaluation expands beyond standard metrics to scrutinize comprehensive response strategies and the alignment between internal reasoning and final answers, offering deep insights into the cognitive consistency of error handling.

\section{Task Formulation and Error Taxonomy}
Let the input for a Multimodal Large Language Model (MLLM) be a triplet $I$, defined as:
\begin{equation}
I = \{T, V, Q\}
\end{equation}
where $T$ represents the \textbf{Textual Context} (core concepts and technical terms), $V$ denotes the \textbf{Visual Information} (images or charts providing spatial features), and $Q$ specifies the \textbf{Query} (the task instruction to be answered). 
Here, $T$ and $V$ together constitute the \textbf{Premises} serving as the evidential basis for reasoning, while $Q$ defines the task \textbf{Goal} that the model is expected to fulfill.

\begin{figure*}[t]%
    \centering
    \includegraphics[width=0.9\textwidth]{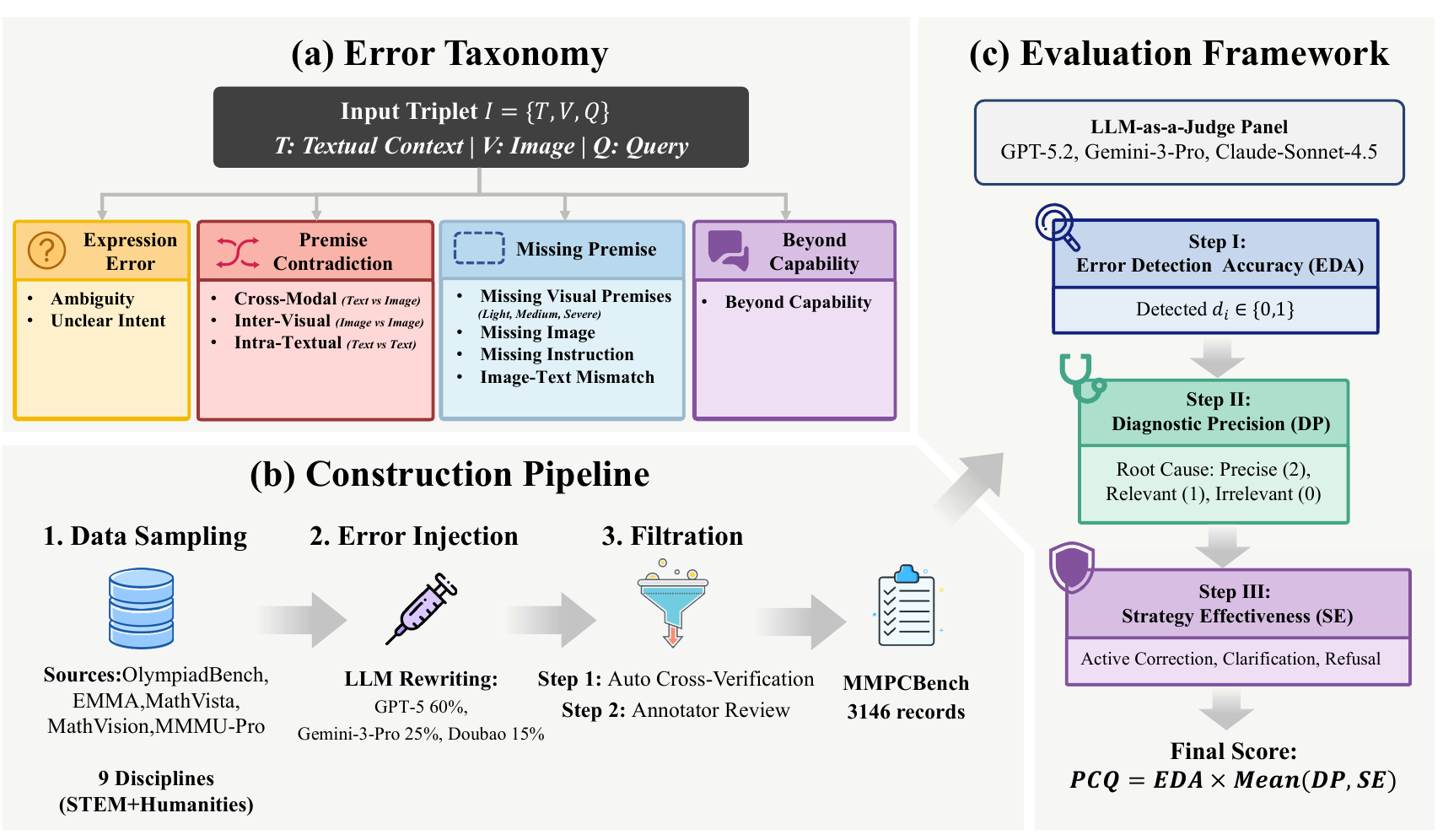}
    \caption{
    \textbf{The Evaluation Pipeline.} 
    \textbf{(a) Error Taxonomy:} Classification of input defects into Expression Error, Premise Contradiction, Missing Premise, and Beyond Capability. 
    \textbf{(b) Construction Pipeline:} A three-stage process featuring multi-source data sampling, multi-model error injection (12k samples), and automated/human filtration. 
    \textbf{(c) Evaluation Framework:} Hierarchical assessment via \textbf{EDA}, \textbf{DP}, and \textbf{SE} metrics, integrated into the composite \textbf{PCQ} score using an LLM-as-a-Judge panel.
    }
    \label{fig:Main}
\end{figure*}

When the input $I$ contains inherent defects, we categorize such defects into \textbf{four primary error types comprising 12 subcategories in total}, classified by the nature of the defect across the input components $T$, $V$, and $Q$. The overview of our error taxonomy, dataset construction pipeline, and the hierarchical evaluation framework is illustrated in Fig.~\ref{fig:Main}.

\paragraph{Expression Error (2 subcategories)} refers to linguistic defects within $T$ or $Q$ that preclude a definitive semantic interpretation. Specifically, \textit{Ambiguity} occurs when textual descriptions of necessary visual premises are too vague, involving unclear references or uncertain scopes, which prevents the model from converging on a single valid interpretation. Meanwhile, \textit{Unclear Intent} denotes instances where the instruction $Q$ is either incomplete or overly abstract, making the intended task goal unidentifiable.

\paragraph{Premise Contradiction (3 subcategories)} arises when the background premises provided in input $I$ contain logically mutually exclusive information. This category manifests as \textit{Cross-Modal Contradiction}, defined as a factual conflict between text $T$ and image $V$ regarding essential attributes; \textit{Inter-Visual Contradiction}, where multiple images within $V$ present conflicting evidence for the same visual premise; and \textit{Intra-Textual Contradiction}, which involves internal logical inconsistencies within $T$ that cannot simultaneously be true.

\paragraph{Missing Premise (6 subcategories)} signifies that essential 
information required to solve the task is absent from the input. 
When visual cues are lost due to occlusion or damage, we distinguish 
three extent-based subcategories: \textit{Light Missing Visual Premises} 
at $\approx$30\%, \textit{Medium Missing Visual Premises} at $\approx$50\%, 
and \textit{Severe Missing Visual Premises} at $\approx$80\%. 
This category further encompasses \textit{Missing Image}, where no visual 
input is provided; \textit{Missing Instruction}, where the task query is 
absent; and \textit{Mismatch}, where the semantic content of $V$ is 
irrelevant to the domain of $T$ and $Q$.

\paragraph{Beyond Capability (1 subcategory)} characterizes tasks where 
the goal $Q$ requires answers that cannot be logically inferred from the 
available evidence in $V$ and $T$. Specifically, \textit{Beyond Capability} 
covers cases such as predicting inherently unknowable futures or reasoning 
about non-existent entity attributes, rendering the query objectively 
unanswerable regardless of the model's reasoning capabilities.

\section{Benchmark Construction}

The construction of \textbf{MMPCBench} proceeds in three stages: data sampling, error injection, and filtration. We initially collect samples from existing datasets. These samples are then modified by Large Language Models (LLMs) to introduce specific errors. Finally, the data undergoes a two-step validation process involving both model-based verification and human review to determine the final set of unanswerable samples.

\subsection{Data Sampling}
We compiled a source pool from datasets such as OlympiadBench~\cite{he2024olympiadbench}, MMMU-Pro~\cite{yue2025mmmu}, EMMA~\cite{hao2025can}, MathVista~\cite{lu2023mathvista}, and MATH-Vision~\cite{wang2024measuring}.
The selection covers nine disciplines, ranging from STEM fields (e.g., Physics, Engineering) to Humanities and Social Sciences.
Based on these sources, we modified the original questions and contexts to construct new samples, ensuring the data differs from the original benchmarks while maintaining the underlying subject matter.

\subsection{Error Injection}
To inject specific error types, we utilized three models: GPT-5.1 (high)~\cite{openai2025gpt51}, Gemini-3-Pro~\cite{gemini3pro}, and Doubao-Seed-1.6-vision~\cite{bytedanceseed2025} to rewrite the samples.
The injection process applies specific strategies for different error categories:

\paragraph{Visual Defects:} We apply gradient-guided degradation to remove key semantic regions (e.g., edges) for \textit{Missing Visual Premise}, and utilize generative editing tools (e.g., Gemini-3-Pro-Image~\cite{gemini3image}) to synthesize \textit{Inter-Visual Contradictions}.

\paragraph{Textual and Logic Defects:} For errors such as \textit{Ambiguity} and \textit{Unclear Intent}, we employ few-shot prompting to guide the models in rewriting queries or inserting contradictory premises into the text.

\paragraph{Structural Defects:} We perform batch processing to shuffle or remove data components, creating \textit{Mismatch} or \textit{Missing Modality} scenarios.

Through this process, a total of 12k candidate samples were synthesized, where per-category injection strategies and prompt templates are detailed in Appendix~\ref{sec:error-injection}.

\subsection{Filtration and Quality Assurance}
To ensure data quality, we applied a two-step screening process evaluating each sample on two criteria: \textit{error-type compliance} (the injected flaw conforms to its target category definition) and \textit{question unanswerability} (the modified question is objectively unanswerable).

In Stage 1, each sample was independently verified by the two models that did not generate it using binary pass/fail judgments; failing either check reduced the 12k candidates to 4,832. In Stage 2, three graduate-student annotators reviewed the remaining samples, with each sample assessed by two annotators and disagreements resolved by a third (Cohen's $\kappa = 0.85$), yielding a final set of 3,146 high-quality instances. Detailed statistics are provided in Appendix~\ref{sec:Appendix Dataset Statistics}.

\section{Evaluation Framework and Metrics}
\label{sec:metrics}

To rigorously assess the proactive critique ability of MLLMs, specifically their proficiency in identifying, diagnosing, and resolving erroneous inputs, we establish a hierarchical evaluation framework comprising three sequential metrics: \textit{Error Detection Accuracy}, \textit{Diagnostic Precision}, and \textit{Strategic Effectiveness}. Let $\mathcal{D} = \{x_i\}_{i=1}^N$ denote the set consisting of $N$ erroneous inputs. The evaluation proceeds as follows:

\subsection{Step I: Error Detection Accuracy (EDA).}
This step measures the model's fundamental capability to proactively identify the presence of an error. For the $i$-th sample, we define a binary variable $d_i \in \{0, 1\}$: $d_i=1$ (Detected) is assigned \textbf{only if} the response contains an explicit negative statement clearly pointing out that the input is unreasonable or erroneous; $d_i=0$ (Missed) covers complying with the erroneous instruction, hallucinated affirmative responses, \textbf{or silent corrections without explicit acknowledgment}. The EDA metric is:
\begin{equation}
    \text{EDA} = \frac{1}{N} \sum_{i=1}^{N} d_i
\end{equation}
We denote the set of successfully detected samples as $\mathcal{H} = \{i \mid d_i = 1\}$, which serves as the evaluation domain for Steps II and III.

\subsection{Step II: Diagnostic Precision (DP).}
This step assesses the quality of diagnosis regarding the error's root cause, evaluated \textbf{only} for $i \in \mathcal{H}$. We assign $s_i^{diag} \in \{0, 1, 2\}$: \textbf{2 (Precise)} if the model accurately pinpoints the core logic trap or factual error consistent with ground truth; \textbf{1 (Relevant)} if the explanation is pertinent but misses the fundamental root cause or contains deviations; \textbf{0 (Irrelevant)} if the explanation is logically unrelated to the actual defect. The DP metric is defined as:
\begin{equation}
    \text{DP} = \frac{1}{|\mathcal{H}|} \sum_{i \in \mathcal{H}} s_i^{diag}
\end{equation}

\subsection{Step III: Strategic Effectiveness (SE).}
This step evaluates the utility and actionability of remedial actions for $i \in \mathcal{H}$. The strategy space comprises three mutually exclusive types: \textit{Correction}, \textit{Clarification}, and \textit{Refusal}. We assign $s_i^{strat} \in \{0, 1, 2\}$ \textbf{based on the explicitness and utility of the response}: \textbf{2} for an effective response (e.g., correct answer under explicit assumptions or specific actionable guidance); \textbf{1} for a weak attempt (e.g., vague clarification or insufficient guidance); \textbf{0} if no valid strategy was employed. The strategy-distribution figure additionally uses \textit{None} for responses with no detected critique and \textit{Unsure} when the judges do not agree on a strategy. The SE metric is defined as:
\begin{equation}
    \text{SE} = \frac{1}{|\mathcal{H}|} \sum_{i \in \mathcal{H}} s_i^{strat}
\end{equation}

\paragraph{Proactive Critique Quality (PCQ).}
To provide a holistic assessment, PCQ integrates the global detection rate with the quality of diagnosis and strategy, serving as an expected measure of overall performance:
\begin{equation}
    \text{PCQ} = \text{EDA} \times \frac{\text{DP}/2 + \text{SE}/2}{2}
\end{equation}
where DP and SE are normalized to $[0, 1]$ by dividing by their maximum score of 2.

\paragraph{Automated Evaluation Protocol}
We adopt the LLM-as-a-Judge paradigm with a panel of three advanced models: GPT-5.2~\cite{openai2025gpt52}, Gemini-3-Pro~\cite{gemini3pro}, and Claude-Sonnet-4~\cite{anthropic2025claude4sonnet}, applying metric-specific aggregation protocols. For EDA, we use majority voting (positive if $\geq$2 judges agree). For DP, we average scores from all three judges. For SE, we first identify the dominant strategy type via majority vote and average scores from the aligned judges. In cases of complete disagreement, the strategy is designated as ''Unsure'' and the final score is averaged across all three judges.

\paragraph{Evaluation Reliability}
We validate our LLM-as-a-Judge protocol with both inter-judge and human-alignment analyses. Across the full benchmark, the three-judge panel achieves Fleiss' $\kappa=0.70$. On 180 stratified samples annotated by three graduate annotators, the overall Human--LLM agreement is $\kappa=0.71$ with 78\% accuracy, supporting the reliability of our protocol for large-scale evaluation. More details are provided in Appendix~\ref{sec:human_reliability}.

\begin{table*}[t]%
    \centering
    \resizebox{\textwidth}{!}{%
    \begin{tabular}{l|cccc|cccc}
    \toprule
    \multirow{2}{*}{\textbf{Models}} & \multicolumn{4}{c|}{\textbf{Overall Performance}} & \multicolumn{4}{c}{\textbf{Category-wise PCQ (Score)}} \\
    \cmidrule(lr){2-5} \cmidrule(lr){6-9}
     & \textbf{EDA} (\%) & \textbf{DP} (0-2) & \textbf{SE} (0-2) & \textbf{PCQ} & \textbf{Expr.} & \textbf{Contr.} & \textbf{Miss.} & \textbf{Beyond} \\
    \midrule
    \multicolumn{9}{l}{\textit{\textbf{Closed-Source Models}}} \\
    \midrule
    Claude-Sonnet-4.5 (Thinking) & 40.2 & 1.66 & 1.86 & 35.4 & 13.8 & 17.9 & \textbf{44.6} & 85.4 \\
    Gemini-3-Pro         & 37.4 & 1.46 & 1.79 & 30.4 & 13.6 & 39.2 & 24.9 & 64.2 \\
    Gemini-3-Flash       & 32.3 & 1.51 & 1.77 & 26.5 & 10.4 & 35.7 & 19.9 & 62.8 \\
    GPT-5.1 (high)                 & 31.7 & \textbf{1.79} & \textbf{1.95} & 29.6 & 14.4 & 7.2 & 37.6 & \underline{87.5} \\
    GPT-5.2                      & 28.2 & 1.66 & 1.87 & 24.9 & 15.3 & 10.7 & 25.8 & 81.3 \\
    GPT-5-mini                   & 24.8 & \underline{1.75} & \underline{1.93} & 22.9 & 12.9 & 6.6 & 28.1 & 65.0 \\
    Doubao-Seed-1.6-vision            & 14.9 & 1.23 & 1.57 & 10.4 & 2.9 & 10.6 & 6.7 & 42.6 \\
    \midrule
    \multicolumn{9}{l}{\textit{\textbf{Open-Source Models}}} \\
    \midrule
    Qwen3-VL-235B-A22B-Instruct       & \textbf{60.0} & 1.47 & 1.69 & \textbf{47.4} & \textbf{30.7} & \textbf{51.3} & \underline{42.2} & \textbf{93.9} \\
    Qwen3-VL-30B-A3B-Instruct        & \underline{55.0} & 1.37 & 1.57 & \underline{40.4} & \underline{24.8} & \underline{40.5} & 38.3 & 80.7 \\
    Qwen3-VL-8B-Instruct         & 53.3 & 1.36 & 1.54 & 38.7 & 21.3 & 37.9 & 36.8 & 84.0 \\
    Gemma-3-12B-it               & 27.0 & 1.27 & 1.42 & 18.1 & 6.4 & 11.1 & 21.0 & 48.2 \\
    Gemma-3-27B-it               & 25.9 & 1.30 & 1.42 & 17.6 & 5.8 & 11.9 & 20.6 & 43.4 \\
    Gemma-3-4B-it                & 18.6 & 0.91 & 1.24 & 10.0 & 5.1 & 6.0 & 11.8 & 22.4 \\
    GLM-4.6V                     & 16.0 & 1.51 & 1.69 & 12.8 & 2.6 & 6.9 & 11.9 & 54.6 \\
    \midrule
    \textit{Average}             & \textit{33.2} & \textit{1.45} & \textit{1.67} & \textit{26.1} & \textit{12.9} & \textit{21.0} & \textit{26.4} & \textit{65.4} \\
    \bottomrule
    \end{tabular}%
    }
    \caption{
    \textbf{Main Results of Proactive Critique Ability.} 
    Evaluation of 14 MLLMs. 
    Metrics include \textbf{Error Detection Accuracy (EDA, \%)}, \textbf{Diagnostic Precision (DP, 0--2)}, \textbf{Strategic Effectiveness (SE, 0--2)}, and the composite \textbf{Proactive Critique Quality (PCQ, 0--100)}.
    The right block details PCQ scores across four error categories: \textit{Expression Error (Expr.)}, \textit{Premise Contradiction (Contr.)}, \textit{Missing Premise (Miss.)}, and \textit{Beyond Capability (Beyond)}.
    \textbf{Bold} denotes the best result globally, and \underline{underlined} denotes the second best.
    The bottom row shows the average performance across all evaluated models.
    }
    \label{tab:main_results}
\end{table*}%

\section{Experiment}

\subsection{Setup}

We assess a range of large language and multimodal large language models, including both closed-source and open-source systems. For closed-source models, we evaluate Gemini-3-Pro~\cite{gemini3pro}, GPT-5.1 (high)~\cite{openai2025gpt51}, Claude-Sonnet-4.5 (Thinking)~\cite{anthropic2025claude45sonnet}, Doubao-Seed-1.6-vision~\cite{bytedanceseed2025}, Gemini-3-Flash~\cite{gemini3flash}, GPT-5-mini~\cite{openai2025gpt5}, GPT-5.2~\cite{openai2025gpt52}. For open-source models, we evaluate Qwen3-VL-235B-A22B-Instruct~\cite{qwen3vlreport}, Qwen3-VL-30B-A3B-Instruct~\cite{qwen3vlreport}, Qwen3-VL-8B-Instruct~\cite{qwen3vlreport}, GLM-4.6V~\cite{glm46v}, Gemma-3-27B-it~\cite{gemma3_27b_it}, Gemma-3-12B-it~\cite{gemma3_12b_it}, Gemma-3-4B-it~\cite{gemma3_4b_it}.

Detailed model descriptions and hyperparameter configurations are provided in Appendix~\ref{sec:Detailed setup}.

\subsection{Main Result}
\label{sec:main_results}

\subsubsection*{Proactive Critique Is Widely Deficient}
\label{sec:model_comparison}
As shown in Table~\ref{tab:main_results}, proactive critique remains a systematic weakness across current MLLMs: the average PCQ is only 26.1 and the average EDA is 33.2\%, indicating that models fail to produce any critical response in approximately two-thirds of erroneous-input scenarios, instead defaulting to sycophantic task completion.

The performance distribution follows a clear three-tier hierarchy: the Qwen3-VL family leads; mainstream proprietary models (Claude, Gemini-3 series, and the GPT series) form a middle tier; while the Gemma-3 series, GLM-4.6V, and Doubao-Seed-1.6-vision constitute the bottom tier.

The distribution further exhibits pronounced within-family clustering: intra-family PCQ gaps across model scales remain below 10 points, whereas cross-family gaps can exceed 30 points. This suggests that proactive critique ability is primarily driven by each family's shared training paradigm rather than model scale.

Notably, scaling benefits are conditional on a capability-supportive training paradigm. Within Qwen3-VL, PCQ improves consistently with scale, from 38.7 (8B) to 47.4 (235B). In contrast, Gemma-3 shows near-stagnation or even slight regression.

\subsubsection*{Proactivity Is the Primary Bottleneck}
\label{sec:detection_vs_reasoning}
The cross-model distribution reveals that PCQ variability is dominated by EDA rather than critique quality (DP/SE). EDA shows far greater inter-model variability (CV\,=\,0.43) than DP (CV\,=\,0.16) and SE (CV\,=\,0.13). Pearson correlations with PCQ further confirm this: EDA is strongly correlated ($r = 0.97$), whereas the quality term $Q = \frac{\mathcal{N}(\mathrm{DP}) + \mathcal{N}(\mathrm{SE})}{2}$ shows only a weak correlation ($r = 0.39$), indicating that cross-model divergence is primarily driven by error detection proactivity rather than diagnostic or remedial quality.

Two distinct behavioral patterns emerge. Leading proprietary models such as GPT-5.1 (high) and Claude-Sonnet-4.5 (Thinking) demonstrate notably higher critique quality upon error detection, with GPT-5.1 (high) achieving DP of 1.79 and SE of 1.95, yet their conservative proactivity (EDA 31.7\% and 40.2\%, respectively) substantially suppresses overall PCQ. In contrast, the Qwen3-VL series achieves its PCQ advantage primarily through higher EDA (53.3\%--60.0\%), despite modestly lower DP (1.36--1.47) and SE (1.54--1.69).

These results establish proactivity as the primary bottleneck: leading proprietary models already possess strong latent critique potential, but their overall PCQ is mainly constrained by conservative error detection.

\subsubsection*{Category-wise Analysis}
\label{sec:category_analysis}

As shown in Table~\ref{tab:main_results}, the overall average PCQ across the four error categories forms a clear hierarchy: Expression Error (12.9), Premise Contradiction (21.0), Missing Premise (26.4), and Beyond Capability (65.4). This reflects the semantic salience of each error type: Beyond Capability errors directly obstruct task completion and are thus readily discoverable, whereas Expression and Contradiction errors require models to proactively scrutinize subtle defects in seemingly valid inputs.

Figure~\ref{fig:category_modality} groups subcategories by the modality carrying the primary error signal. Visual-only errors remain a difficult modality region, with PCQ below 25\% for most model--subcategory pairs. The Light-to-Severe scores do not follow one universal monotonic pattern, showing that nominal degradation severity alone does not determine performance. Cross-modal subcategories exhibit the largest inter-model divergence, as they jointly demand text comprehension, visual perception, and cross-modal consistency verification, making them a highly discriminative modality dimension.

\subsection{The Consistency Gap Between Reasoning and Response}
\label{sec:alignment}

\begin{table}[t]%
\centering
\resizebox{\linewidth}{!}{%
\begin{tabular}{l c c c c}
\toprule
\textbf{Model}
& \textbf{EDA}$^{R}$ ($\uparrow$)
& \textbf{EDA}$^{F}$ ($\uparrow$)
& \textbf{DDR} ($\downarrow$)
& \textbf{DQD} ($\downarrow$) \\
\midrule
Claude-Sonnet-4.5 (Thinking)  & \textbf{0.51} & \textbf{0.40} & \textbf{0.28} & \textbf{0.11} \\
Gemini-3-Pro       & 0.51          & 0.37          & 0.34          & 0.17          \\
Gemini-3-Flash     & 0.47          & 0.32          & 0.37          & 0.17          \\
GLM-4.6V           & 0.42          & 0.16          & 0.63          & 0.19          \\
Doubao-Seed-1.6-vision  & 0.44          & 0.15          & 0.68          & 0.37          \\
\midrule
\textit{Average}               & 0.47          & 0.28          & 0.46          & 0.20          \\
\textit{Coefficient of Variation}                 & 0.08          & 0.38          & 0.35          & 0.43          \\
\bottomrule
\end{tabular}%
}
\caption{\textbf{Consistency Analysis between Reasoning and Response.} $\text{EDA}^{R}$ and $\text{EDA}^{F}$ denote Error Detection Accuracy on the reasoning chain and final response, respectively. \textbf{DDR} (Detection Drop Rate): proportion of reasoning-detected errors absent from the final response (lower is better). \textbf{DQD} (Diagnostic Quality Drop): average decline in diagnostic precision when both components detect the error (lower is better).}
\label{tab:alignment_single_col}
\end{table}%

To evaluate consistency between a model's internal reasoning ($R$) and final response ($F$), we assess error detection on both components using $\text{EDA}^R$/$\text{EDA}^F$ and $\text{DP}^R$/$\text{DP}^F$, and introduce two metrics to quantify critique degradation during the transition.

The \textbf{Detection Drop Rate (DDR)} measures the fraction of reasoning-detected errors absent from the final response. Let $\mathcal{H}_R = \{i \mid d_i^R = 1\}$; DDR is defined as:
\begin{equation}
    \text{DDR} = \frac{1}{|\mathcal{H}_R|} \sum_{i \in \mathcal{H}_R} \mathbb{I}(d_i^F = 0)
\end{equation}
The \textbf{Diagnostic Quality Drop (DQD)} captures a subtler degradation: even when both $R$ and $F$ detect the error, diagnostic precision may decline. Let $\mathcal{H}_{RF} = \{i \mid d_i^R = 1,\ d_i^F = 1\}$; DQD is:
\begin{equation}
    \text{DQD} = \frac{1}{|\mathcal{H}_{RF}|} \sum_{i \in \mathcal{H}_{RF}} \left(s_i^R - s_i^F\right)
\end{equation}
where $s_i^R, s_i^F \in [0, 2]$ are Diagnostic Precision scores in $R$ and $F$, respectively.

Table~\ref{tab:alignment_single_col} reveals a systematic gap across all five models: $\text{EDA}^R$ exceeds $\text{EDA}^F$ by 19 percentage points on average, and DDR shows that 46\% of reasoning-level critique findings are absent from the final response. Two tiers emerge: Claude-Sonnet-4.5 (Thinking), Gemini-3-Pro, and Gemini-3-Flash retain 63\%--72\% of detections (DDR 0.28--0.37), while GLM-4.6V and Doubao-Seed-1.6-vision discard over 60\% (DDR 0.63--0.68). Crucially, $\text{EDA}^R$ is tightly clustered across models (CV\,=\,0.08), whereas $\text{EDA}^F$ is far more dispersed (CV\,=\,0.38), indicating that performance divergence stems not from reasoning-stage detection ability, but from how faithfully those findings are preserved in the output.

DQD reveals a second, milder layer of degradation: on $\mathcal{H}_{RF}$, diagnostic precision still drops by a mean of 0.20 out of 2. This is considerably less severe than the broad critique loss captured by DDR. The former reflects partial quality reduction on surfaced critiques, while the latter reflects their complete absence. Together, these results establish the reasoning-to-response consistency gap as a central bottleneck limiting the proactive critique capability of current reasoning models.

\subsection{Strategy Distribution}
\begin{figure}[t] 
    \centering
    \includegraphics[width=1.0\linewidth]{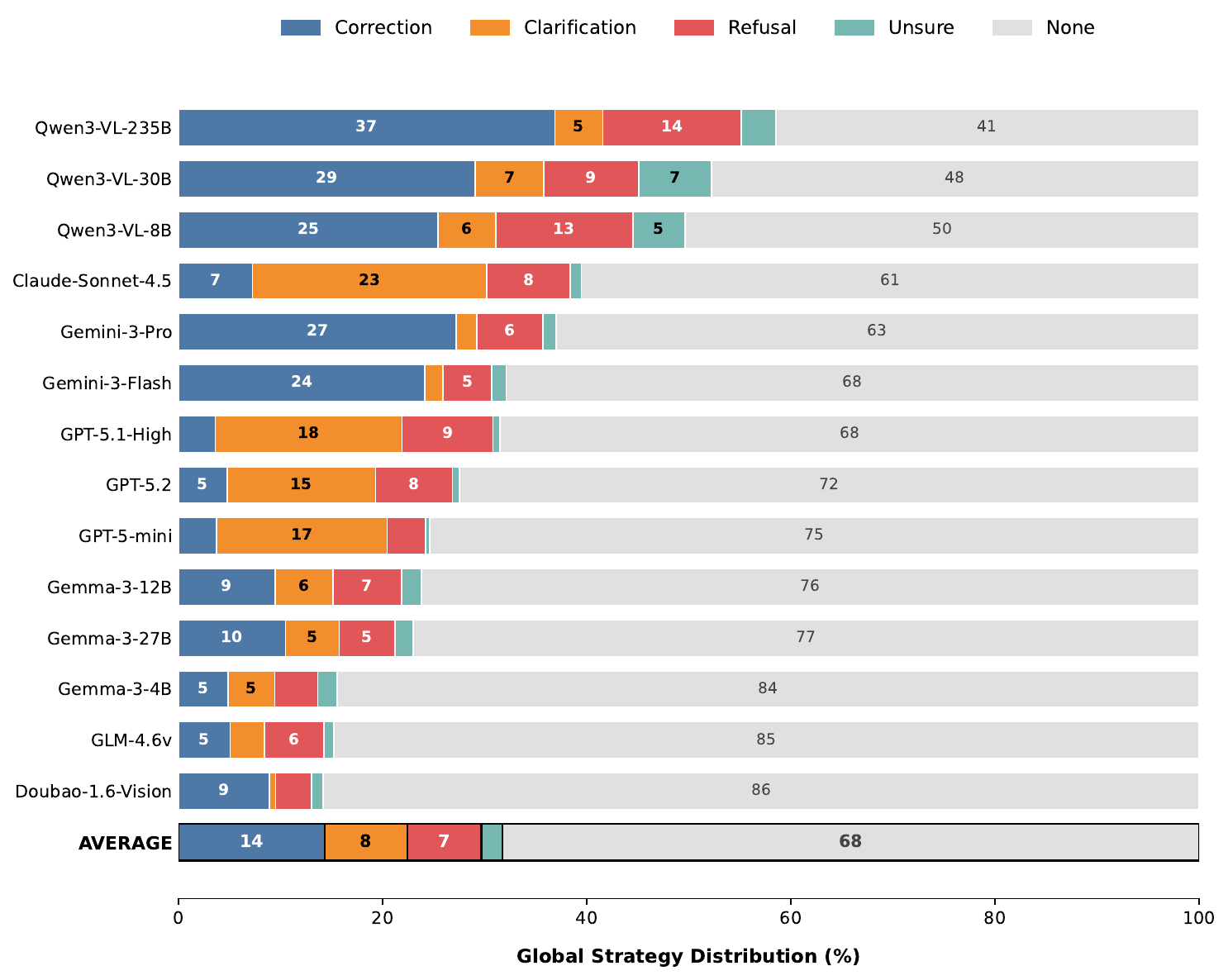}
    
    \caption{
        \textbf{Distribution of response strategies across 14 MLLMs.} 
        The chart illustrates the breakdown of Correction (Blue), Clarification (Orange), Refusal (Red), Unsure (Teal), and None (Grey).
        Values under 4.5\% are omitted for visual clarity.
    }
    \label{fig:strategy_dist_wide}
\end{figure}

Figure~\ref{fig:strategy_dist_wide} illustrates the distribution of response strategies across 14 MLLMs. A pervasive phenomenon is the dominance of the ``None'' strategy, averaging 68\% globally, where models fail to critique flawed inputs and instead yield sycophantic, hallucinated answers. 

Among the models that successfully detect errors, a distinct behavioral bifurcation emerges between aggressive ``Solvers'' and cautious ``Consultants.'' The Qwen3-VL series exemplifies the former approach, prioritizing Correction to maximize immediate utility. Notably, Qwen3-VL-235B-A22B-Instruct achieves a 37\% correction rate—nearly triple the cohort average of 14\%—by autonomously inferring user intent under declared assumptions. This strategy treats visual-textual discrepancies as oversights to be ``repaired'' under explicit assumptions. In contrast, proprietary models function primarily as ``Consultants'' that favor Clarification. Claude-Sonnet-4.5 (Thinking) leads this category with a 23\% clarification rate, followed closely by GPT-5.1 (high) at 18\%. These models guide users to resolve ambiguities explicitly rather than risking unverified guesses.

\section{Conclusion}

We introduce \textbf{MMPCBench}, a comprehensive framework for evaluating Proactive Critique in MLLMs across 3,146 samples and 12 fine-grained error types. Evaluating 14 MLLMs, we find that proactive critique remains a systematic weakness, driven by a pervasive sycophancy that prioritizes instruction-following over input scrutiny. We uncover a consistency gap where reasoning models detect errors internally yet suppress these findings in final responses, identify visual-only errors as a universal blind spot, and observe a behavioral bifurcation between aggressive ``Solvers'' and cautious ``Consultants.'' These findings call for a paradigm shift toward autonomous input validation, offering a clear roadmap for more honest and robust multimodal systems.

\section*{Acknowledgments}
The authors would like to thank the anonymous reviewers for their valuable comments. This work is supported by the National Key Research and Development Program of China (No.2023YFF0905400), the National Natural Science Foundation of China (No.U2341229) and the Reform Commission Foundation of Jilin Province (No.2024C003). 

\section*{Limitation}

We have systematically assessed the proactive critique performance of 14 representative multimodal large language models (MLLMs), yet several limitations of this work need to be acknowledged.

To begin with, our benchmark is limited to English and Chinese text inputs. Given the distinct syntactic features and cultural reasoning patterns of other languages, our findings may not be fully generalizable across linguistic families. Expanding the benchmark to a multilingual setting will be an important focus of our future research.

\bibliography{custom}

\clearpage
\appendix

\section{Details of Experimental Setup}
\label{sec:Detailed setup}

We evaluate a range of closed-source and open-source models. Proprietary models, including GPT-5.2, Claude-Sonnet-4.5 (Thinking), and Gemini-3-Pro, are deployed via their latest official versions using the default settings of their official interfaces. Open-source counterparts are accessed through Hugging Face with hyperparameters configured according to official technical reports. Original source images are provided without additional resizing or preprocessing; injected samples may contain generated, degraded, or replaced images as described below. Image resolution follows each model's default API configuration, reflecting standard deployment conditions. Specifically, the Qwen3-VL series uses $T=0.7$, $top\_p=0.7$, and a presence penalty of $1.5$; GLM-4.6V adopts $T=0.7$ and $top\_p=0.7$; and the Gemma-3 series utilizes $T=1.0$ and $top\_p=0.7$. Detailed model specifications are provided in Table~\ref{tab:model_link}.

\begin{table}[t]%
\centering
\small
\begin{tabular}{lr}
\toprule
\textbf{Error Category} & \textbf{\# of Instance} \\
\midrule
Expression Error & 584 \\
\hspace{2mm}- Ambiguity & 294 \\
\hspace{2mm}- Unclear Intent & 290 \\
\midrule
Premise Contradiction & 862 \\
\hspace{2mm}- Cross-Modal Contradiction & 284 \\
\hspace{2mm}- Inter-Visual Contradiction & 288 \\
\hspace{2mm}- Intra-Textual Contradiction & 290 \\
\midrule
Missing Premise & 1,405 \\
\hspace{2mm}- Missing Visual Premise & 505 \\
\hspace{4mm}- Light & 150 \\
\hspace{4mm}- Medium & 173 \\
\hspace{4mm}- Severe & 182 \\
\hspace{2mm}- Missing Image & 300 \\
\hspace{2mm}- Missing Instruction & 300 \\
\hspace{2mm}- Mismatch & 300 \\
\midrule
Beyond Capability & 295 \\
\hspace{2mm}- Beyond Capability & 295 \\
\midrule
\textbf{Subject Distribution} & \\
\hspace{2mm}- Maths & 622 \\
\hspace{2mm}- Chemistry & 490 \\
\hspace{2mm}- Physics & 417 \\
\hspace{2mm}- Engineering & 354 \\
\hspace{2mm}- CS & 312 \\
\hspace{2mm}- Biology & 283 \\
\hspace{2mm}- Medicine & 282 \\
\hspace{2mm}- Finance & 195 \\
\hspace{2mm}- Society & 191 \\
\midrule
\textbf{Total Instances} & \textbf{3,146}\\
\midrule
Avg. \# of Tokens in Questions & 173.33 \\
Max. \# of Tokens in Questions & 4,359 \\
Min. \# of Tokens in Questions & 5 \\
\midrule
Avg. \# of Images per Sample & 1.16 \\
Max. \# of Images per Sample & 5 \\
Min. \# of Images per Sample & 0 \\
\bottomrule
\end{tabular}%
\caption{Dataset composition and coverage. Counts report the number of benchmark instances by primary error type, fine-grained subtype, and subject/domain.}
\label{tab:dataset-stats}
\end{table}%

\begin{table}[t]%
\centering
\small
\begin{tabular}{lccc}
\toprule
 & \textbf{Human} & \textbf{H-LLM} & \textbf{H-LLM} \\
\textbf{Metric} & Fleiss' $\kappa$ & Cohen's $\kappa$ & Acc. \\
\midrule
EDA     & 0.79 & 0.77 & 88\% \\
DP      & 0.75 & 0.71 & 75\% \\
SE      & 0.72 & 0.67 & 73\% \\
\midrule
Overall & \textbf{0.76} & \textbf{0.71} & \textbf{78\%} \\
\bottomrule
\end{tabular}%
\caption{Per-metric agreement between human annotators and the LLM-as-a-Judge panel on 180 stratified samples. Human Fleiss' $\kappa$ measures inter-annotator agreement; H-LLM Cohen's $\kappa$ and H-LLM Acc. measure agreement and accuracy against the majority-vote human consensus.}
\label{tab:human_reliability}
\end{table}%

\section{Details of Dataset Statistics} 
\label{sec:Appendix Dataset Statistics}

In this section, we provide a comprehensive breakdown of the dataset's composition. To facilitate a deeper understanding of the challenges posed by our benchmark, we categorize the data across three dimensions: error taxonomy, subject diversity, and sample complexity.

\subsection{Error Taxonomy and Distribution}
As summarized in Table~\ref{tab:dataset-stats}, the dataset contains 3,146 instances meticulously categorized into four primary error types. The largest category, \textit{Missing Premise} (1,405 instances), is further divided into fine-grained sub-categories, including varying degrees of visual information deficiency (Light, Medium, and Severe). The subcategory counts provide approximately balanced coverage within several groups (e.g., approximately 290 instances per contradiction subtype), while the overall category sizes reflect the construction targets of the benchmark.

\subsection{Domain Diversity and Complexity}
The dataset spans 9 distinct academic and professional subjects, with a strong emphasis on STEM fields such as Mathematics, Chemistry, and Physics. Furthermore, the dataset exhibits significant complexity in terms of input length and multi-modal integration. The question length varies from a few tokens to over 4,000 tokens, challenging the long-context processing capabilities of Multimodal Large Language Models (MLLMs). Samples contain between 0 and 5 images; multi-image subsets additionally require inter-visual reasoning.

\section{Error Injection and Construction Details}
\label{sec:error-injection}

\paragraph{LLM-based Rewriting.}
This strategy covers five error types: \textit{Ambiguity}, \textit{Unclear Intent}, \textit{Cross-Modal Contradiction}, \textit{Intra-Textual Contradiction}, and \textit{Beyond Capability}.
For each type, a single LLM (GPT-5.1 (high) for 60\% of samples, Gemini-3-Pro for 25\%, and Doubao-Seed-1.6-vision for 15\%) receives the original multimodal question and is prompted via few-shot instructions to rewrite the text so as to introduce the target error type, while leaving the image unchanged.
Injection prompts for each of these five types are shown in Figures~\ref{fig:Ambiguity}--\ref{fig:Unclear_Intent}.

\paragraph{Inter-Visual Contradiction.}
This type requires single-image samples only. Construction proceeds in two steps: (1) an LLM generates a targeted image-editing instruction describing how to alter a key visual premise in the original image; (2) Gemini-3-Pro-Image~\cite{gemini3image} synthesizes a second image according to this instruction. The original and newly generated images are then presented together, so that the two images provide contradictory evidence for a necessary visual premise. The injection prompt is shown in Figure~\ref{fig:Inter_Visual_Contradiction}.

\paragraph{Gradient-Guided Visual Degradation.}
This strategy handles three severity levels of \textit{Missing Visual Premise}: Light, Medium, and Severe.
We apply a gradient-guided attack that identifies high-salience edge regions in the original image and applies a degradation operation (blurring, pixelation, or geometric warping) within those regions. The proportion of the salient region degraded is $20$--$30\%$ for \textit{Light}, $40$--$50\%$ for \textit{Medium}, and $70$--$80\%$ for \textit{Severe}, ensuring the target visual premise is reliably obscured at each level.

\paragraph{Complete Modality Removal.}
This strategy covers two types: \textit{Missing Image} and \textit{Missing Instruction}.
The former is constructed by removing all images from the question, retaining only the text; the latter by removing all text and retaining only the images. Neither operation requires LLM involvement.

\paragraph{Text-Image Mismatch.}
A mismatched sample is constructed by replacing the original images with semantically unrelated images drawn from a different subject domain in the dataset pool. The number of replacement images matches the original, preserving question structure while making the visual content entirely irrelevant.

\section{Details of Prompt Templates}

Prompts corresponding to Figures~\ref{fig:Error_Detection_Accuracy}--\ref{fig:Strategic_Effectiveness} are employed to evaluate Error Detection Accuracy, Diagnostic Precision, and Strategic Effectiveness. In contrast, prompts associated with Figures~\ref{fig:Ambiguity}--\ref{fig:Unclear_Intent} are used to synthesize evaluation samples spanning six categories: Ambiguity, Beyond Capability, Inter-Visual Contradiction, Intra-Textual Contradiction, Cross-Modal Contradiction, and Unclear Intent.

\section{Representative Samples}
Table~\ref{tab:Sample_1} presents representative examples of Ambiguity, Beyond Capability, and Intra-Textual Contradiction error types in our dataset, where question--image pairs are unanswerable due to linguistic vagueness, unmet reasoning requirements, or internal logical inconsistencies.
Table~\ref{tab:Sample_2} illustrates Inter-Visual Contradiction, Missing Instruction, and Missing Image, highlighting cases in which unanswerability arises from cross-image conflicts or missing critical information.
Table~\ref{tab:Sample_3} provides qualitative examples of Missing Visual Premise, grouped by the severity of information loss (Light, Medium, Severe), demonstrating how visual occlusion or corruption progressively undermines answerability.
Finally, Table~\ref{tab:Sample_4} showcases Cross-Modal Contradiction, Mismatch, and Unclear Intent, where inconsistencies between textual and visual modalities or ill-defined task objectives preclude valid responses.

\section{Supplementary Evaluation Results}
\subsection{Bootstrap Confidence Intervals for PCQ}
\label{sec:Appendix Significance}

Table~\ref{tab:pcq_ci} reports PCQ scores with 95\% bootstrap confidence intervals (CI) estimated from $B=1000$ resamplings of the evaluation set ($N=3146$), with adjacent-rank comparisons based on paired-bootstrap p-values.

The majority of adjacent-rank gaps are statistically significant ($p < 0.05$). The separation at the top of the ranking is particularly pronounced: Qwen3-VL-235B-A22B-Instruct (PCQ~47.4, CI~[45.9,~48.9]) and Claude-Sonnet-4.5 (Thinking) (PCQ~35.4, CI~[33.6,~37.0]) have non-overlapping intervals. Four adjacent pairs yield $p \ge 0.05$: Gemini-3-Pro vs.\ GPT-5.1 (high) ($p=0.37$), Gemini-3-Flash vs.\ GPT-5.2 ($p=0.07$), Gemma-3-12B-it vs.\ Gemma-3-27B-it ($p=0.37$), and Doubao-Seed-1.6-vision vs.\ Gemma-3-4B-it ($p=0.43$); these pairs have point-estimate differences of at most 1.6 PCQ points. These bootstrap results support the tier structure reported in the main text.

\begin{table}[t]%
\centering
\setlength{\tabcolsep}{5pt}
\small
\resizebox{\linewidth}{!}{%
\begin{tabular}{lccc}
\toprule
\textbf{Model} & \textbf{PCQ} & \textbf{95\% CI} & \textbf{\textit{Sig.}} \\
\midrule
Qwen3-VL-235B-A22B-Instruct & 47.4 & [45.9, 48.9] & $^{*}$ \\
Qwen3-VL-30B-A3B-Instruct & 40.4 & [38.9, 41.9] & $^{*}$ \\
Qwen3-VL-8B-Instruct & 38.7 & [37.1, 40.2] & $^{*}$ \\
Claude-Sonnet-4.5 (Thinking) & 35.4 & [33.6, 37.0] & $^{*}$ \\
Gemini-3-Pro      & 30.4 & [28.9, 31.9] & n.s.\ ($p=0.37$) \\
GPT-5.1 (high)    & 29.6 & [28.1, 31.2] & $^{*}$ \\
Gemini-3-Flash    & 26.5 & [25.1, 27.8] & n.s.\ ($p=0.07$) \\
GPT-5.2           & 24.9 & [23.4, 26.4] & $^{*}$ \\
GPT-5-mini        & 22.9 & [21.5, 24.4] & $^{*}$ \\
Gemma-3-12B-it       & 18.1 & [16.9, 19.2] & n.s.\ ($p=0.37$) \\
Gemma-3-27B-it       & 17.6 & [16.4, 18.8] & $^{*}$ \\
GLM-4.6V          & 12.8 & [11.7, 13.9] & $^{*}$ \\
Doubao-Seed-1.6-vision   & 10.4 & [9.5,  11.4] & n.s.\ ($p=0.43$) \\
Gemma-3-4B-it        & 10.0 & [9.1,  10.8] & {---} \\
\bottomrule
\end{tabular}%
}
\caption{PCQ scores with 95\% bootstrap confidence intervals ($B=1000$).
The \textit{Sig.}\ column reports pairwise bootstrap significance
between each model and the next-ranked model:
$^{*}\,p<0.05$ (significant gap); n.s.\ $p\ge 0.05$ (non-significant gap);
{---} for the last-ranked model.}
\label{tab:pcq_ci}
\vspace{2pt}
\end{table}%

\subsection{Human Validation of the Evaluation Protocol}
\label{sec:human_reliability}

We conduct additional validation to assess the reliability of our LLM-as-a-Judge evaluation protocol. The validation consists of two parts: measuring agreement among the automated judges on the full benchmark, and comparing aggregated LLM judgments with human annotations on a stratified subset.

For automated evaluation, each model response is scored by three advanced LLM judges: GPT-5.2, Gemini-3-Pro, and Claude-Sonnet-4. Across the full benchmark, the three-judge panel achieves Fleiss' $\kappa=0.70$, indicating substantial inter-judge agreement. This result suggests that the evaluation rubric produces stable judgments across different judge models.

We further conduct a human validation study on a stratified subset of 180 questions covering all subtypes. Each sample is independently annotated by three graduate annotators using the same rubric as the automated judges. Human inter-annotator agreement reaches Fleiss' $\kappa=0.79$ for EDA, $0.75$ for DP, and $0.72$ for SE, with an overall value of $0.76$.

Using the majority-vote human consensus as the reference, the aggregated LLM judgment achieves Cohen's $\kappa=0.77$ for EDA, $0.71$ for DP, and $0.67$ for SE. The corresponding accuracies are 88\%, 75\%, and 73\%, respectively, with an overall $\kappa=0.71$ and overall accuracy of 78\%. These results are summarized in Table~\ref{tab:human_reliability} and show substantial agreement between automated and human judgments. 

Overall, both automated inter-judge agreement and Human--LLM agreement exceed the commonly used substantial-agreement threshold of $\kappa>0.61$~\cite{landis1977measurement}, indicating that our automated protocol is reliable for large-scale evaluation on MMPCBench.

\subsection{Fine-Grained Evaluation Results}

The EDA results in Table \ref{tab:eda} underscore a substantial performance disparity between proprietary and open-source models in multimodal reasoning. Notably, the Qwen3-VL series leads across many dimensions, with its 235B variant approaching ceiling performance in Intra-Textual and Mismatch detection. While the Gemini-3 series demonstrates specialized proficiency in identifying Cross-Modal contradictions, Inter-Visual inconsistencies remain a pervasive bottleneck across the evaluated models, including GPT-5. This gap indicates that contemporary Multimodal Large Language Models (MLLMs) are more reliable at identifying explicit text-image discrepancies than at resolving fine-grained logical conflicts within the visual modality.

Diagnostic Precision (DP) metrics (Table \ref{tab:dp}) reveal a persistent decoupling between coarse-grained recognition and precise logical reasoning. Beyond Capability and Mismatch are relatively easier for many models, whereas precision is generally lower for Inter-Visual Contradiction and Missing Visual Premise. GPT-5.1 (high) and Claude-Sonnet-4.5 (Thinking) remain among the stronger models on several visual-severity conditions, but the Light-to-Severe pattern varies across model families rather than following a universal monotonic trend. Identifying the existence of an error is therefore easier than providing a high-fidelity diagnosis for complex cross-modal inconsistencies.

The Strategic Effectiveness (SE) results (Table \ref{tab:se}) show that GPT-5-mini is near-optimal across most visual and textual sub-categories, while GPT-5.1 (high) exhibits a pronounced exception on Inter-Visual Contradiction. Gemini-3-Pro and Claude-Sonnet-4.5 (Thinking) remain strong on selected Cross-Modal and Expression Error tasks, with model-specific variation across visual-missing severities. The low Inter-Visual scores of GLM-4.6V and the Gemma-3 series indicate that detecting a discrepancy does not guarantee an effective strategy for reconciling conflicting visual data.

The PCQ evaluation (Table \ref{tab:pcq}) exposes a significant bottleneck in autonomous error detection, with models performing substantially lower than in assisted diagnostic settings. Qwen3-VL-235B-A22B-Instruct retains leadership in Intra-Textual and Mismatch scenarios, whereas Inter-Visual Contradiction and Missing Visual Premise remain difficult across most models. The Gemini-3 series exhibits competitive proactive sensitivity in Cross-Modal tasks but performs poorly on Missing Instruction compared with the GPT and Claude families. These findings identify subtle, localized visual-logical flaws as a central challenge for self-correcting multimodal architectures.

\section{Additional Analyses}
\subsection{Prompt Sensitivity Analysis}
\label{app:prompt_sensitivity}

To verify that our main experimental findings are not artifacts of a particular prompt wording, we conduct a prompt sensitivity analysis on a stratified subset of 300 samples drawn from the full test set (12 subcategories $\times$ 25 samples, seed\,=\,42). We design three semantically equivalent prompt variants: \textit{orig.}\ (identical to the main experiment), \textit{formal}\ (rewritten in a formal academic register), and \textit{concise}\ (reduced to a minimal direct instruction). Crucially, all three variants are strictly prohibited from containing any language that hints at the presence of errors, preserving the proactive nature of the critique task. We evaluate three representative models and report EDA and PCQ scores in Table~\ref{tab:prompt_robustness}.

Results show that performance is stable across prompt variants. The maximum absolute shift in EDA across all models and variants is 2.7 percentage points (Qwen3-VL-235B-A22B-Instruct, \textit{concise}), and the maximum PCQ shift is 2.9 points (Gemini-3-Pro, \textit{concise}). Model rankings remain consistent across prompt styles.

\begin{table}[h]%
\centering
\resizebox{\linewidth}{!}{%
\small
\begin{tabular}{llcccc}
\toprule
\textbf{Model} & \textbf{Prompt} & \textbf{EDA(\%)} & \textbf{PCQ} & \textbf{$\Delta$EDA} & \textbf{$\Delta$PCQ} \\
\midrule
\multirow{3}{*}{Qwen3-VL-235B-A22B-Instruct}
  & orig.   & 55.3 & 37.0 & \phantom{$-$}0.0 & \phantom{$-$}0.0 \\
  & formal  & 55.3 & 39.2 & \phantom{$-$}0.0 & \phantom{$-$}2.2 \\
  & concise & 52.7 & 37.6 &          $-$2.7  & \phantom{$-$}0.5 \\
\midrule
\multirow{3}{*}{GPT-5-mini}
  & orig.   & 18.3 & 17.5 & \phantom{$-$}0.0 & \phantom{$-$}0.0 \\
  & formal  & 20.0 & 18.2 & \phantom{$-$}1.7 & \phantom{$-$}0.7 \\
  & concise & 19.7 & 16.5 & \phantom{$-$}1.3 &          $-$1.0 \\
\midrule
\multirow{3}{*}{Gemini-3-Pro}
  & orig.   & 30.7 & 20.6 & \phantom{$-$}0.0 & \phantom{$-$}0.0 \\
  & formal  & 31.3 & 22.7 & \phantom{$-$}0.7 & \phantom{$-$}2.1 \\
  & concise & 29.3 & 23.5 &          $-$1.3  & \phantom{$-$}2.9 \\
\bottomrule
\end{tabular}%
}
\caption{Prompt sensitivity of three representative models on a stratified 300 sample subset.
\textit{Orig.}\ replicates the main-experiment prompt; \textit{formal} rewrites it in an
academic register; \textit{concise} reduces it to a minimal direct instruction.
$\Delta$EDA and $\Delta$PCQ are absolute differences (in percentage points) relative to
\textit{orig.}\ per model.}
\label{tab:prompt_robustness}
\end{table}%

\subsection{Reasoning Effort and Budget Ablation}
\label{app:reasoning_budget}

We analyze reasoning-response consistency for Claude-Sonnet-4.5 (Thinking), whose interface exposes stable reasoning content. We sample 10 flawed inputs from each of the 12 error subcategories (120 inputs in total) and vary two request parameters independently: \texttt{reasoning\_effort} $\in$ \{low, medium, high\} and \texttt{max\_tokens} $\in$ \{4,096, 8,192, 16,384\}. No additional system prompt or error hint is added. Reasoning and final responses are evaluated separately with the same rubric.

\begin{table}[t]%
\centering\small\setlength{\tabcolsep}{4pt}
\resizebox{\linewidth}{!}{%
\begin{tabular}{lcccc}\toprule
\textbf{Reasoning effort} & $\mathrm{EDA}^{R}$ & $\mathrm{EDA}^{F}$ & $\mathrm{DDR}$ & $\mathrm{DQD}$ \\
\midrule
Low & 47.7\% & 39.3\% & 27.0\% & -0.10 \\
Medium & 53.0\% & 41.3\% & 25.7\% & 0.02 \\
High & \textbf{53.7\%} & \textbf{46.7\%} & \textbf{19.0\%} & 0.04 \\
\bottomrule\end{tabular}
}
\caption{Reasoning-effort ablation for Claude-Sonnet-4.5 (Thinking), averaged over three maximum-token budgets and 120 samples. $\mathrm{EDA}^{R}$ and $\mathrm{EDA}^{F}$ are reasoning- and final-response Error Detection Accuracy; DDR and DQD are Detection Drop Rate and Diagnostic Quality Drop.}
\label{tab:reasoning_budget_active}
\end{table}%

Table~\ref{tab:reasoning_budget_active} reports the effort-level averages for Claude-Sonnet-4.5 (Thinking). Increasing effort from low to high raises $\mathrm{EDA}^{R}$ from 47.7\% to 53.7\% and $\mathrm{EDA}^{F}$ from 39.3\% to 46.7\%, while reducing DDR from 27.0\% to 19.0\%. The corresponding budget-specific results at high effort are shown in Table~\ref{tab:reasoning_budget_consistency}.

\begin{table}[t]%
\centering\small\setlength{\tabcolsep}{4pt}
\resizebox{\linewidth}{!}{%
\begin{tabular}{lccccc}\toprule
\textbf{Reasoning effort} & \textbf{Maximum budget} & $\mathrm{EDA}^{R}$ & $\mathrm{EDA}^{F}$ & $\mathrm{DDR}$ & $\mathrm{DQD}$ \\
\midrule
High & 4,096 & 54\% & \textbf{48\%} & \textbf{18\%} & 0.06 \\
High & 8,192 & 54\% & 45\% & 21\% & 0.00 \\
High & 16,384 & 53\% & 47\% & \textbf{18\%} & 0.06 \\
\bottomrule\end{tabular}
}
\caption{Reasoning-budget ablation for Claude-Sonnet-4.5 (Thinking) at high effort on 120 samples. $\mathrm{EDA}^{R}$ and $\mathrm{EDA}^{F}$ are reasoning- and final-response Error Detection Accuracy; DDR and DQD are Detection Drop Rate and Diagnostic Quality Drop.}
\label{tab:reasoning_budget_consistency}
\end{table}%

At high effort, increasing the maximum budget does not yield a monotonic gain: $\mathrm{EDA}^{F}$ is 48\%, 45\%, and 47\%, while DDR is 18\%, 21\%, and 18\% across the three budgets. DQD remains close to zero.

\subsection{Valid-Input Split: False Alarms and Precision}
\label{app:negative_split}

As the main benchmark only contains erroneous inputs ($\text{H}_{pos}$), we additionally evaluate valid inputs ($\text{H}_{neg}$) to quantify false alarms and over-criticism. We use the original benchmark items as candidates for the valid condition, but do not assume that every source item is intrinsically valid: some items labeled as correct in the source data still contain ambiguities, contradictions, or other defects. We therefore apply an independent validity check and retain an item only when both validators judge the unmodified question and image to be valid. This filtering leaves 639 valid/error pairs. The valid condition uses the unmodified question and image, whereas the paired error condition uses the corresponding injected-error item. The paired analysis below covers the seven models for which complete valid-input results are available; error-condition detection scores are reused from the main evaluation protocol.

For each paired item, we report EDA/Recall, the false-alarm rate (FAR), Precision, and F1. FAR is the proportion of valid inputs incorrectly flagged as defective; Precision is the proportion of flagged inputs that are actually defective; and F1 is the harmonic mean of Recall and Precision. We additionally report PCQ-F1, computed as $\mathrm{F1}\times((\mathrm{DP}/2+\mathrm{SE}/2)/2)$. Table~\ref{tab:valid_false_alarm} reports these metrics for the seven models with complete valid-input results.

\begin{table}[t]%
\centering
\small
\setlength{\tabcolsep}{4pt}
\resizebox{\linewidth}{!}{%
\begin{tabular}{lccccc}
\toprule
\textbf{Model} & \textbf{EDA/Recall} & \textbf{FAR} & \textbf{Precision} & \textbf{F1} & \textbf{PCQ-F1} \\
\midrule
Qwen3-VL-235B-A22B-Instruct & 58.8 & 18.8 & 75.8 & 66.3 & 52.1 \\
Claude-Sonnet-4.5 (Thinking) & 39.6 & 1.56 & 96.2 & 56.1 & 49.5 \\
Qwen3-VL-30B-A3B-Instruct & 55.7 & 20.3 & 73.3 & 63.3 & 47.6 \\
Gemini-3-Pro & 38.3 & 1.25 & 96.8 & 54.9 & 45.3 \\
Qwen3-VL-8B-Instruct & 52.6 & 21.8 & 70.7 & 60.3 & 44.2 \\
Gemini-3-Flash & 34.3 & 2.03 & 94.4 & 50.3 & 41.7 \\
GPT-5-mini & 25.2 & 0.94 & 96.4 & 40.0 & 37.2 \\
\bottomrule
\end{tabular}%
}
\caption{False-alarm analysis on the filtered valid/error paired split. EDA/Recall is Error Detection Accuracy on error inputs; FAR, Precision, F1, and PCQ-F1 denote False-Alarm Rate, precision, F1, and the PCQ-weighted F1 score.}
\label{tab:valid_false_alarm}
\end{table}%

The corresponding valid-input URR-V values are reported in Table~\ref{tab:valid_urr}. URR-V never exceeds 0.16\%, with no model producing more than one unfounded full refusal among the 639 valid inputs.

\begin{table}[t]%
\centering
\small
\setlength{\tabcolsep}{5pt}
\resizebox{\linewidth}{!}{%
\begin{tabular}{lc}
\toprule
\textbf{Model} & \textbf{Valid-input URR-V (\%) $\downarrow$} \\
\midrule
Qwen3-VL-235B-A22B-Instruct & 0.16 \\
Qwen3-VL-30B-A3B-Instruct & 0.16 \\
Qwen3-VL-8B-Instruct & 0.00 \\
Claude-Sonnet-4.5 (Thinking) & 0.00 \\
Gemini-3-Pro & 0.00 \\
Gemini-3-Flash & 0.00 \\
GPT-5-mini & 0.00 \\
\bottomrule
\end{tabular}%
}
\caption{Unfounded refusal rate on the 639-item valid-input split (URR-V). The table reports the percentage of valid inputs that received a full refusal without an underlying error.}
\label{tab:valid_urr}
\end{table}%

These results provide an auxiliary view of the proactivity--over-criticism trade-off: models with higher error recall can also produce more false alarms on valid inputs, whereas conservative models obtain higher precision. We therefore use this split as a robustness analysis rather than replacing the main error-only PCQ evaluation.

\subsection{Refusal Rates and Unfounded Refusals}
\label{app:refusal_rates}

Figure~3 shows the strategy distribution; the exact refusal rates on flawed inputs are given in Table~\ref{tab:refusal_rate}. They range from 3.50\% to 13.57\%, indicating model-specific remediation preferences rather than an all-or-nothing behavior. These rates should be interpreted together with URR-V, which measures unsupported refusal on valid inputs.

\begin{table}[t]%
\centering
\small
\setlength{\tabcolsep}{4pt}
\resizebox{\linewidth}{!}{%
\begin{tabular}{lc}
\toprule
\textbf{Model} & \textbf{Flawed-input Refusal (\%)} \\
\midrule
Qwen3-VL-235B-A22B-Instruct & 13.57 \\
Qwen3-VL-8B-Instruct & 13.41 \\
Qwen3-VL-30B-A3B-Instruct & 9.28 \\
GPT-5.1 (high) & 8.90 \\
Claude-Sonnet-4.5 (Thinking) & 8.17 \\
GPT-5.2 & 7.53 \\
Gemma-3-12B-it & 6.74 \\
Gemini-3-Pro & 6.45 \\
GLM-4.6V & 5.82 \\
Gemma-3-27B-it & 5.47 \\
Gemini-3-Flash & 4.77 \\
Gemma-3-4B-it & 4.23 \\
GPT-5-mini & 3.75 \\
Doubao-Seed-1.6-vision & 3.50 \\
\bottomrule
\end{tabular}%
}
\caption{Full-refusal rates on the 3,146 flawed inputs. Values denote the proportion of responses assigned the refusal strategy.}
\label{tab:refusal_rate}
\end{table}%

\subsection{Judge Strategy Bias and Model-Family Robustness}
\label{app:judge_robustness}

Table~\ref{tab:strategy_bias} reports mean SE by response strategy for each judge. Clarification and Refusal differ by only 0.03, 0.01, and 0.05 for GPT-5.2, Gemini-3-Pro, and Claude-Sonnet-4, respectively. The human validation gives 73\% agreement for SE ($\kappa=0.67$), so the evidence does not indicate a specific Clarification-over-Refusal preference.

The reported response-model--judge-model DP values are shown in Table~\ref{tab:judge_matrix_dp}. The three GPT models and Claude-Sonnet-4.5 (Thinking) receive their highest reported DP from the Gemini judge rather than their same-family judge. The leave-family-out ranking remains highly consistent with the original (Spearman $\rho=0.969$; Kendall $\tau=0.912$).

\begin{table}[!t]%
\centering
\small
\setlength{\tabcolsep}{6pt}
\resizebox{\linewidth}{!}{%
\begin{tabular}{lccc}
\toprule
\textbf{Judge} & \textbf{Correction} & \textbf{Clarification} & \textbf{Refusal} \\
\midrule
GPT-5.2 & 1.28 & 1.90 & 1.87 \\
Gemini-3-Pro & 1.79 & 1.99 & 1.98 \\
Claude-Sonnet-4 & 1.66 & 1.93 & 1.88 \\
\bottomrule
\end{tabular}%
}
\caption{Mean Strategic Effectiveness (SE) scores assigned to clarification, correction, and refusal responses by the three LLM judges.}
\label{tab:strategy_bias}
\end{table}%

\begin{table}[!t]%
\centering
\small
\setlength{\tabcolsep}{5pt}
\resizebox{\linewidth}{!}{%
\begin{tabular}{lccc}
\toprule
\textbf{Response model} & \textbf{GPT judge} & \textbf{Gemini judge} & \textbf{Claude judge} \\
\midrule
GPT-5.1 (high) & 1.718 & 1.873 & 1.562 \\
GPT-5-mini & 1.683 & 1.825 & 1.451 \\
GPT-5.2 & 1.620 & 1.770 & 1.472 \\
Gemini-3-Pro & 1.565 & 1.708 & 1.199 \\
Gemini-3-Flash & 1.601 & 1.755 & 1.245 \\
Claude-Sonnet-4.5 (Thinking) & 1.548 & 1.820 & 1.544 \\
\bottomrule
\end{tabular}%
}
\caption{Judge robustness for Diagnostic Precision (DP). Rows identify the model that generated the response and columns identify the judge model; entries are mean DP scores on the reported evaluation subset (0--2 scale).}
\label{tab:judge_matrix_dp}
\end{table}%

\begin{table}[!htbp]%
\centering
\small
\setlength{\tabcolsep}{3pt}
\resizebox{\linewidth}{!}{%
\begin{tabular}{lccccccc}
\toprule
\textbf{Model} & \textbf{MEDA} & \textbf{DP-cov.} & \textbf{SE} & \textbf{PCQ} & \textbf{Type cov.} & \textbf{All det.} & \textbf{Partial det.} \\
\midrule
Qwen3-VL-8B-Instruct & 76 & 1.114 & 1.105 & 42.2 & 44 & 16 & 72 \\
Qwen3-VL-30B-A3B-Instruct & 74 & 1.157 & 0.889 & 37.6 & 46 & 27 & 65 \\
Qwen3-VL-235B-A22B-Instruct & 74 & 1.090 & 0.811 & 35.2 & 42 & 16 & 68 \\
GPT-5-mini & 46 & 1.087 & 1.261 & 27.0 & 25 & 8 & 42 \\
Gemini-3-Flash & 30 & 1.178 & 1.200 & 17.8 & 19 & 10 & 28 \\
\bottomrule
\end{tabular}%
}
\caption{Results on the 50-sample dual-error subset. MEDA denotes Multi-Error Detection Accuracy; DP-cov., Diagnostic-Point Coverage; SE, Strategic Effectiveness; and PCQ, Proactive Critique Quality. Type cov. is average ground-truth error-type coverage; All det. and Partial det. indicate detection of both and at least one ground-truth error type, respectively.}
\label{tab:multi_error}
\end{table}%

\subsection{Evaluation on Interacting Errors}

We evaluated 50 dual-error samples covering three interacting-error combinations: Ambiguity + Missing Visual Premise (17), Cross-Modal Contradiction + Unclear Intent (17), and Missing Instruction + Text--Image Mismatch (16). MEDA measures detection of at least one defect; DP-coverage measures diagnostic coverage on the original 0--2 scale; Type Coverage, All Detected, and Partial Detected measure error-type coverage, detection of both types, and detection of at least one ground-truth type, respectively. The results are reported in Table~\ref{tab:multi_error}.

The Qwen3-VL models reach MEDA values of 74\%--76\%, while the best Type Coverage is 46\% and the best All Detected rate is 27\%. Models therefore more often recognize that a compound input is problematic than identify all interacting defects.

\section{Qualitative Cases}

Figure \ref{fig:Non_Reasoning_Failed} shows a failure case of GPT-5.2 in detecting query ambiguity. Conversely, Figure \ref{fig:Non_Reasoning_Success} showcases the robustness of GPT-5-mini in handling samples beyond its capabilities, where it correctly requests missing information. Finally, Figure \ref{fig:Reasoning_No_Aligned} highlights a reasoning-response misalignment in Claude-Sonnet-4.5 (Thinking). Despite the internal Chain-of-Thought correctly identifying a cross-modal contradiction, the final response ignores this issue, demonstrating a lack of alignment between the model's latent reasoning and its observable output.

\begin{table*}[t]%
\centering
\resizebox{\textwidth}{!}{%
\setlength{\tabcolsep}{3.5pt}
\begin{tabular}{lcccccccccccc}
\toprule
\multirow{2}{*}{\textbf{Model}} & \multicolumn{2}{c}{\textbf{Expression Error}} & \multicolumn{3}{c}{\textbf{Premise Contradiction}} & \multicolumn{6}{c}{\textbf{Missing Premise}} & \multicolumn{1}{c}{\textbf{Beyond Capability}} \\
\cmidrule(lr){2-3} \cmidrule(lr){4-6} \cmidrule(lr){7-12} \cmidrule(lr){13-13}
 & Ambiguity & Unclear Intent & Cross-Modal & Inter-Visual & Intra-Textual & Visual Light & Visual Medium & Visual Severe & Missing Image & Missing Instruction & Mismatch & Beyond Capability \\
\midrule
\textbf{Claude-Sonnet-4.5 (Thinking)} & 15.7 & 21.0 & 21.8 & 3.1 & 41.0 & 8.0 & 11.0 & 19.8 & 70.7 & 69.7 & 74.0 & 87.1 \\
\textbf{Doubao-Seed-1.6-vision} & 5.1 & 5.5 & 17.6 & 2.4 & 24.1 & 9.3 & 7.5 & 5.0 & 9.3 & 6.3 & 26.7 & 50.2 \\
\textbf{GLM-4.6V} & 2.7 & 7.9 & 11.3 & 1.0 & 16.9 & 4.0 & 4.0 & 2.2 & 40.3 & 8.3 & 18.0 & 58.0 \\
\textbf{GPT-5-mini} & 7.1 & 24.1 & 9.5 & 1.0 & 15.9 & 5.3 & 8.7 & 14.8 & 44.7 & 46.3 & 30.3 & 67.5 \\
\textbf{GPT-5.1 (high)} & 5.4 & 28.6 & 10.6 & 0.4 & 17.9 & 12.7 & 8.1 & 15.9 & 47.0 & 63.0 & 53.0 & 89.1 \\
\textbf{GPT-5.2} & 7.5 & 30.7 & 16.2 & 3.5 & 24.8 & 8.7 & 8.7 & 9.9 & 44.7 & 37.0 & 38.0 & 82.4 \\
\textbf{Gemini-3-Flash} & 14.6 & 15.2 & 63.0 & 19.1 & 44.1 & 8.0 & 10.4 & 6.0 & 19.7 & 10.3 & 77.7 & 69.2 \\
\textbf{Gemini-3-Pro} & 19.4 & 19.7 & 70.1 & 20.8 & 49.5 & 22.0 & 17.9 & 15.4 & 28.7 & 9.3 & 83.3 & 69.5 \\
\textbf{Gemma-3-12B-it} & 9.2 & 15.5 & 15.5 & 8.7 & 32.8 & 16.0 & 17.3 & 18.1 & 41.7 & 21.7 & 53.3 & 59.3 \\
\textbf{Gemma-3-27B-it} & 12.6 & 13.1 & 15.1 & 8.7 & 32.4 & 16.7 & 15.6 & 16.5 & 41.3 & 19.7 & 51.0 & 53.9 \\
\textbf{Gemma-3-4B-it} & 8.2 & 15.2 & 11.3 & 8.0 & 17.6 & 14.7 & 11.6 & 11.5 & 37.3 & 20.3 & 24.0 & 34.6 \\
\textbf{Qwen3-VL-235B-A22B-Instruct} & 42.5 & 48.3 & 66.9 & 34.7 & 83.1 & 42.0 & 51.4 & 46.2 & 71.0 & 25.0 & 93.0 & 98.3 \\
\textbf{Qwen3-VL-30B-A3B-Instruct} & 36.0 & 39.0 & 56.7 & 30.2 & 73.5 & 54.0 & 47.4 & 45.6 & 76.3 & 18.0 & 86.3 & 89.1 \\
\textbf{Qwen3-VL-8B-Instruct} & 34.0 & 39.0 & 43.7 & 32.3 & 71.7 & 43.3 & 48.0 & 37.4 & 71.0 & 25.7 & 88.0 & 91.5 \\
\midrule
\textit{Mean} & \textit{15.7} & \textit{23.1} & \textit{30.7} & \textit{12.4} & \textit{39.0} & \textit{18.9} & \textit{19.1} & \textit{18.9} & \textit{46.0} & \textit{27.2} & \textit{56.9} & \textit{71.4} \\
\bottomrule
\end{tabular}%
}
\caption{Error Detection Accuracy (EDA, \%) across fine-grained error subtypes and four primary error categories.}
\label{tab:eda}
\end{table*}%

\begin{table*}[t]%
\centering
\resizebox{\textwidth}{!}{%
\setlength{\tabcolsep}{3.5pt}
\begin{tabular}{lcccccccccccc}
\toprule
\multirow{2}{*}{\textbf{Model}} & \multicolumn{2}{c}{\textbf{Expression Error}} & \multicolumn{3}{c}{\textbf{Premise Contradiction}} & \multicolumn{6}{c}{\textbf{Missing Premise}} & \multicolumn{1}{c}{\textbf{Beyond Capability}} \\
\cmidrule(lr){2-3} \cmidrule(lr){4-6} \cmidrule(lr){7-12} \cmidrule(lr){13-13}
 & Ambiguity & Unclear Intent & Cross-Modal & Inter-Visual & Intra-Textual & Visual Light & Visual Medium & Visual Severe & Missing Image & Missing Instruction & Mismatch & Beyond Capability \\
\midrule
\textbf{Claude-Sonnet-4.5 (Thinking)} & 1.09 & 1.45 & 1.41 & 1.52 & 1.63 & 0.89 & 0.89 & 1.66 & 1.87 & 1.25 & 1.88 & 1.95 \\
\textbf{Doubao-Seed-1.6-vision} & 0.76 & 0.88 & 1.35 & 0.38 & 1.29 & 0.21 & 0.21 & 0.56 & 0.71 & 1.14 & 1.25 & 1.60 \\
\textbf{GLM-4.6V} & 0.67 & 0.68 & 1.38 & 0.11 & 1.32 & 0.39 & 0.10 & 0.08 & 1.72 & 1.11 & 1.39 & 1.85 \\
\textbf{GPT-5-mini} & 1.19 & 1.52 & 1.11 & 1.56 & 1.23 & 1.88 & 1.93 & 1.91 & 1.98 & 1.65 & 1.91 & 1.91 \\
\textbf{GPT-5.1 (high)} & 1.00 & 1.63 & 1.32 & 0.33 & 1.21 & 1.65 & 1.86 & 1.91 & 1.99 & 1.65 & 1.97 & 1.94 \\
\textbf{GPT-5.2} & 1.02 & 1.47 & 1.15 & 1.13 & 1.31 & 0.97 & 0.78 & 1.04 & 1.95 & 1.49 & 1.88 & 1.97 \\
\textbf{Gemini-3-Flash} & 1.04 & 1.14 & 1.75 & 1.55 & 1.47 & 0.28 & 0.24 & 0.36 & 1.44 & 1.39 & 1.55 & 1.75 \\
\textbf{Gemini-3-Pro} & 1.01 & 1.13 & 1.71 & 1.38 & 1.45 & 0.36 & 0.28 & 0.35 & 1.33 & 1.27 & 1.66 & 1.81 \\
\textbf{Gemma-3-12B-it} & 0.59 & 0.95 & 0.96 & 0.73 & 1.27 & 0.40 & 0.71 & 0.40 & 1.59 & 1.12 & 1.53 & 1.57 \\
\textbf{Gemma-3-27B-it} & 0.59 & 0.82 & 1.19 & 0.91 & 1.33 & 0.32 & 0.54 & 0.58 & 1.56 & 1.24 & 1.62 & 1.57 \\
\textbf{Gemma-3-4B-it} & 0.53 & 0.67 & 0.90 & 0.41 & 1.06 & 0.29 & 0.43 & 0.29 & 1.21 & 1.07 & 0.77 & 1.16 \\
\textbf{Qwen3-VL-235B-A22B-Instruct} & 1.06 & 1.15 & 1.63 & 1.27 & 1.74 & 0.60 & 0.69 & 0.85 & 1.51 & 1.13 & 1.81 & 1.89 \\
\textbf{Qwen3-VL-30B-A3B-Instruct} & 1.08 & 1.18 & 1.38 & 1.20 & 1.67 & 0.56 & 0.63 & 0.74 & 1.43 & 0.94 & 1.65 & 1.78 \\
\textbf{Qwen3-VL-8B-Instruct} & 1.00 & 0.88 & 1.48 & 1.16 & 1.63 & 0.60 & 0.63 & 0.66 & 1.45 & 1.00 & 1.70 & 1.80 \\
\midrule
\textit{Mean} & \textit{0.90} & \textit{1.11} & \textit{1.34} & \textit{0.97} & \textit{1.40} & \textit{0.67} & \textit{0.71} & \textit{0.81} & \textit{1.55} & \textit{1.25} & \textit{1.61} & \textit{1.75} \\
\bottomrule
\end{tabular}%
}
\caption{Diagnostic Precision (DP) scores across fine-grained error subtypes and four primary error categories (0--2 scale).}
\label{tab:dp}
\end{table*}%

\begin{table*}[t]%
\centering
\resizebox{\textwidth}{!}{%
\setlength{\tabcolsep}{3.5pt}
\begin{tabular}{lcccccccccccc}
\toprule
\multirow{2}{*}{\textbf{Model}} & \multicolumn{2}{c}{\textbf{Expression Error}} & \multicolumn{3}{c}{\textbf{Premise Contradiction}} & \multicolumn{6}{c}{\textbf{Missing Premise}} & \multicolumn{1}{c}{\textbf{Beyond Capability}} \\
\cmidrule(lr){2-3} \cmidrule(lr){4-6} \cmidrule(lr){7-12} \cmidrule(lr){13-13}
 & Ambiguity & Unclear Intent & Cross-Modal & Inter-Visual & Intra-Textual & Visual Light & Visual Medium & Visual Severe & Missing Image & Missing Instruction & Mismatch & Beyond Capability \\
\midrule
\textbf{Claude-Sonnet-4.5 (Thinking)} & 1.66 & 1.74 & 1.61 & 1.83 & 1.72 & 1.39 & 1.32 & 1.81 & 1.89 & 1.97 & 1.91 & 1.98 \\
\textbf{Doubao-Seed-1.6-vision} & 1.48 & 1.32 & 1.69 & 1.24 & 1.58 & 1.25 & 0.91 & 1.39 & 1.24 & 1.73 & 1.42 & 1.79 \\
\textbf{GLM-4.6V} & 1.56 & 1.17 & 1.47 & 0.33 & 1.67 & 0.78 & 0.98 & 1.12 & 1.79 & 1.55 & 1.50 & 1.92 \\
\textbf{GPT-5-mini} & 1.75 & 1.91 & 1.80 & 2.00 & 1.75 & 2.00 & 1.98 & 2.00 & 1.99 & 1.94 & 1.98 & 1.95 \\
\textbf{GPT-5.1 (high)} & 1.57 & 1.92 & 1.80 & 0.00 & 1.78 & 1.84 & 2.00 & 1.97 & 1.98 & 1.99 & 1.98 & 1.98 \\
\textbf{GPT-5.2} & 1.71 & 1.86 & 1.63 & 1.58 & 1.65 & 1.44 & 1.19 & 1.41 & 1.97 & 1.95 & 1.95 & 1.98 \\
\textbf{Gemini-3-Flash} & 1.69 & 1.72 & 1.85 & 1.77 & 1.69 & 1.67 & 1.61 & 1.29 & 1.71 & 1.91 & 1.70 & 1.88 \\
\textbf{Gemini-3-Pro} & 1.68 & 1.76 & 1.87 & 1.74 & 1.71 & 1.60 & 1.60 & 1.57 & 1.76 & 1.85 & 1.81 & 1.89 \\
\textbf{Gemma-3-12B-it} & 1.15 & 1.35 & 1.25 & 1.06 & 1.25 & 0.74 & 0.77 & 0.97 & 1.57 & 1.76 & 1.46 & 1.68 \\
\textbf{Gemma-3-27B-it} & 1.07 & 1.14 & 1.24 & 0.97 & 1.41 & 0.89 & 0.63 & 0.84 & 1.57 & 1.81 & 1.53 & 1.65 \\
\textbf{Gemma-3-4B-it} & 0.85 & 1.27 & 0.92 & 0.93 & 1.22 & 0.50 & 1.10 & 0.86 & 1.38 & 1.77 & 1.01 & 1.44 \\
\textbf{Qwen3-VL-235B-A22B-Instruct} & 1.57 & 1.62 & 1.77 & 1.52 & 1.76 & 1.43 & 1.27 & 1.41 & 1.62 & 1.79 & 1.75 & 1.93 \\
\textbf{Qwen3-VL-30B-A3B-Instruct} & 1.44 & 1.57 & 1.57 & 1.43 & 1.59 & 1.27 & 1.48 & 1.36 & 1.59 & 1.44 & 1.58 & 1.84 \\
\textbf{Qwen3-VL-8B-Instruct} & 1.40 & 1.41 & 1.60 & 1.49 & 1.63 & 1.03 & 1.04 & 1.09 & 1.47 & 1.56 & 1.65 & 1.87 \\
\midrule
\textit{Mean} & \textit{1.47} & \textit{1.55} & \textit{1.58} & \textit{1.28} & \textit{1.60} & \textit{1.27} & \textit{1.28} & \textit{1.36} & \textit{1.68} & \textit{1.79} & \textit{1.66} & \textit{1.84} \\
\bottomrule
\end{tabular}%
}
\caption{Strategic Effectiveness (SE) scores across fine-grained error subtypes and primary error categories (0--2 scale).}
\label{tab:se}
\end{table*}%

\begin{table*}[t]%
\centering
\resizebox{\textwidth}{!}{%
\setlength{\tabcolsep}{3.5pt}
\begin{tabular}{lcccccccccccc}
\toprule
\multirow{2}{*}{\textbf{Model}} & \multicolumn{2}{c}{\textbf{Expression Error}} & \multicolumn{3}{c}{\textbf{Premise Contradiction}} & \multicolumn{6}{c}{\textbf{Missing Premise}} & \multicolumn{1}{c}{\textbf{Beyond Capability}} \\
\cmidrule(lr){2-3} \cmidrule(lr){4-6} \cmidrule(lr){7-12} \cmidrule(lr){13-13}
 & Ambiguity & Unclear Intent & Cross-Modal & Inter-Visual & Intra-Textual & Visual Light & Visual Medium & Visual Severe & Missing Image & Missing Instruction & Mismatch & Beyond Capability \\
\midrule
\textbf{Claude-Sonnet-4.5 (Thinking)} & 10.8 & 16.8 & 16.5 & 2.6 & 34.4 & 4.6 & 6.1 & 17.2 & 66.3 & 56.2 & 70.3 & 85.4 \\
\textbf{Doubao-Seed-1.6-vision} & 2.9 & 3.0 & 13.4 & 1.0 & 17.3 & 3.4 & 2.1 & 2.4 & 4.6 & 4.5 & 17.8 & 42.6 \\
\textbf{GLM-4.6V} & 1.5 & 3.7 & 8.0 & 0.1 & 12.6 & 1.2 & 1.1 & 0.7 & 35.4 & 5.5 & 13.0 & 54.6 \\
\textbf{GPT-5-mini} & 5.2 & 20.7 & 6.9 & 0.9 & 11.8 & 5.2 & 8.5 & 14.5 & 44.3 & 41.5 & 29.5 & 65.0 \\
\textbf{GPT-5.1 (high)} & 3.5 & 25.4 & 8.2 & 0.0 & 13.4 & 11.1 & 7.8 & 15.4 & 46.7 & 57.4 & 52.3 & 87.5 \\
\textbf{GPT-5.2} & 5.1 & 25.6 & 11.3 & 2.4 & 18.4 & 5.2 & 4.3 & 6.0 & 43.7 & 31.8 & 36.3 & 81.3 \\
\textbf{Gemini-3-Flash} & 10.0 & 10.8 & 56.6 & 15.8 & 34.9 & 3.9 & 4.8 & 2.5 & 15.5 & 8.5 & 63.1 & 62.8 \\
\textbf{Gemini-3-Pro} & 13.0 & 14.2 & 62.8 & 16.3 & 39.0 & 10.8 & 8.4 & 7.3 & 22.1 & 7.3 & 72.3 & 64.2 \\
\textbf{Gemma-3-12B-it} & 4.0 & 8.9 & 8.6 & 3.9 & 20.6 & 4.6 & 6.4 & 6.2 & 32.9 & 15.6 & 39.8 & 48.2 \\
\textbf{Gemma-3-27B-it} & 5.2 & 6.5 & 9.2 & 4.1 & 22.3 & 5.0 & 4.6 & 5.8 & 32.4 & 15.0 & 40.2 & 43.4 \\
\textbf{Gemma-3-4B-it} & 2.8 & 7.4 & 5.1 & 2.7 & 10.0 & 2.9 & 4.4 & 3.3 & 24.2 & 14.4 & 10.7 & 22.4 \\
\textbf{Qwen3-VL-235B-A22B-Instruct} & 28.0 & 33.4 & 56.9 & 24.2 & 72.8 & 21.3 & 25.1 & 26.2 & 55.5 & 18.3 & 82.9 & 93.9 \\
\textbf{Qwen3-VL-30B-A3B-Instruct} & 22.8 & 26.8 & 41.9 & 19.9 & 59.7 & 24.8 & 25.1 & 23.9 & 57.8 & 10.7 & 69.7 & 80.7 \\
\textbf{Qwen3-VL-8B-Instruct} & 20.4 & 22.2 & 33.7 & 21.4 & 58.4 & 17.6 & 20.1 & 16.4 & 51.8 & 16.4 & 73.8 & 84.0 \\
\midrule
\textit{Mean} & \textit{9.7} & \textit{16.1} & \textit{24.2} & \textit{8.2} & \textit{30.4} & \textit{8.7} & \textit{9.2} & \textit{10.6} & \textit{38.1} & \textit{21.7} & \textit{48.0} & \textit{65.4} \\
\bottomrule
\end{tabular}%
}
\caption{Proactive Critique Quality (PCQ, \%) across fine-grained error subtypes and primary error categories.}
\label{tab:pcq}
\end{table*}%

\clearpage

\begin{figure*}[t]%
  \centering
  \includegraphics[width=\textwidth]{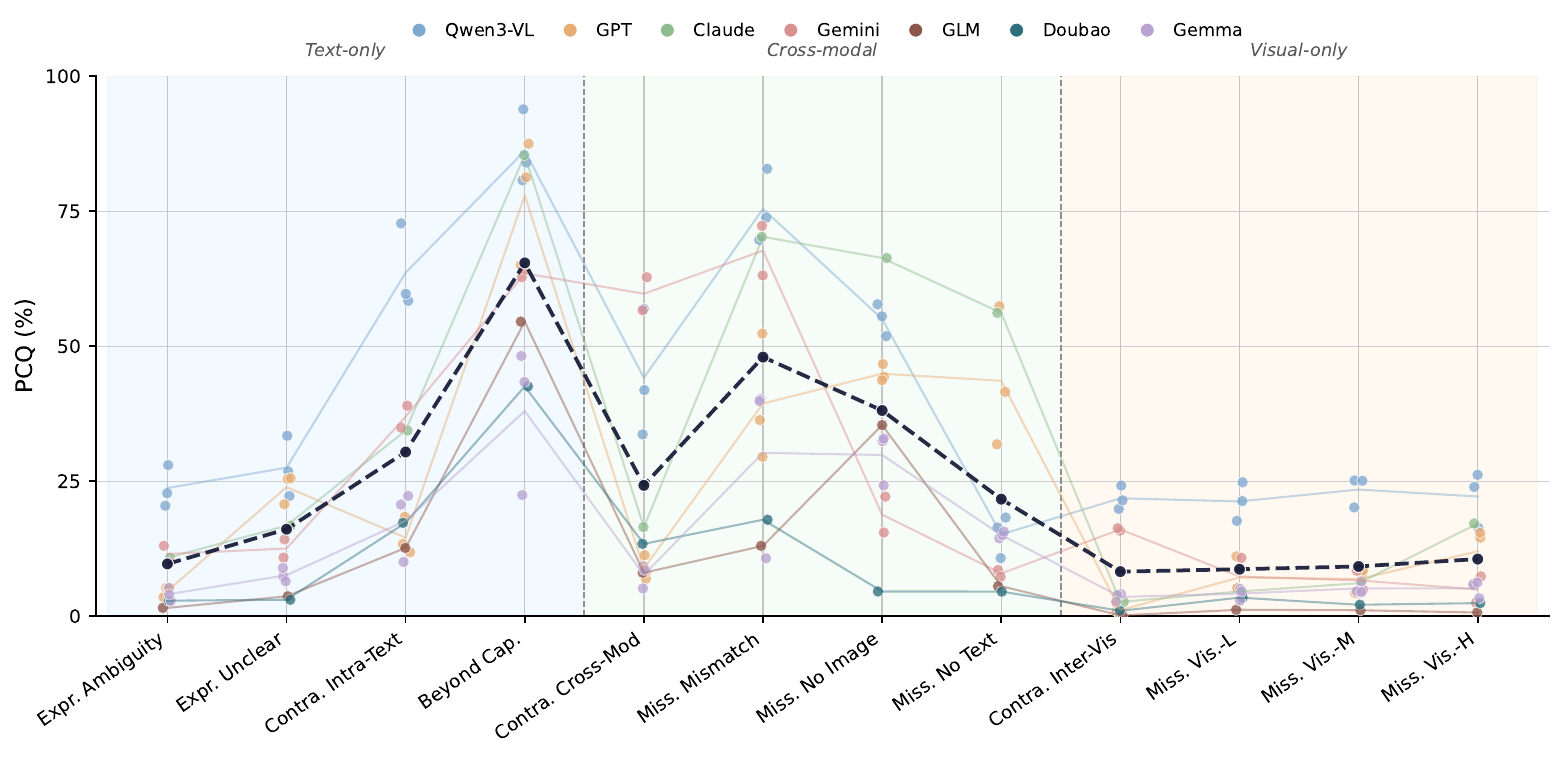}
  \caption{
    Per-subcategory PCQ (\%) across 12 subcategories, grouped into three modality regions based on which modality carries the error signal (Text-only, Cross-modal, Visual-only). Each colored line connects the within-family average across subcategories, and the black dashed line shows the overall average across all 14 models. Subcategory names follow the format "Primary Error Type. Subcategory Name" (e.g., Expr. Ambiguity for Expression Error, Ambiguity subcategory).
  }
  \label{fig:category_modality}
\end{figure*}

\clearpage

\begin{figure*}[!t]%
    \centering

    \includegraphics[width=1.0\textwidth]{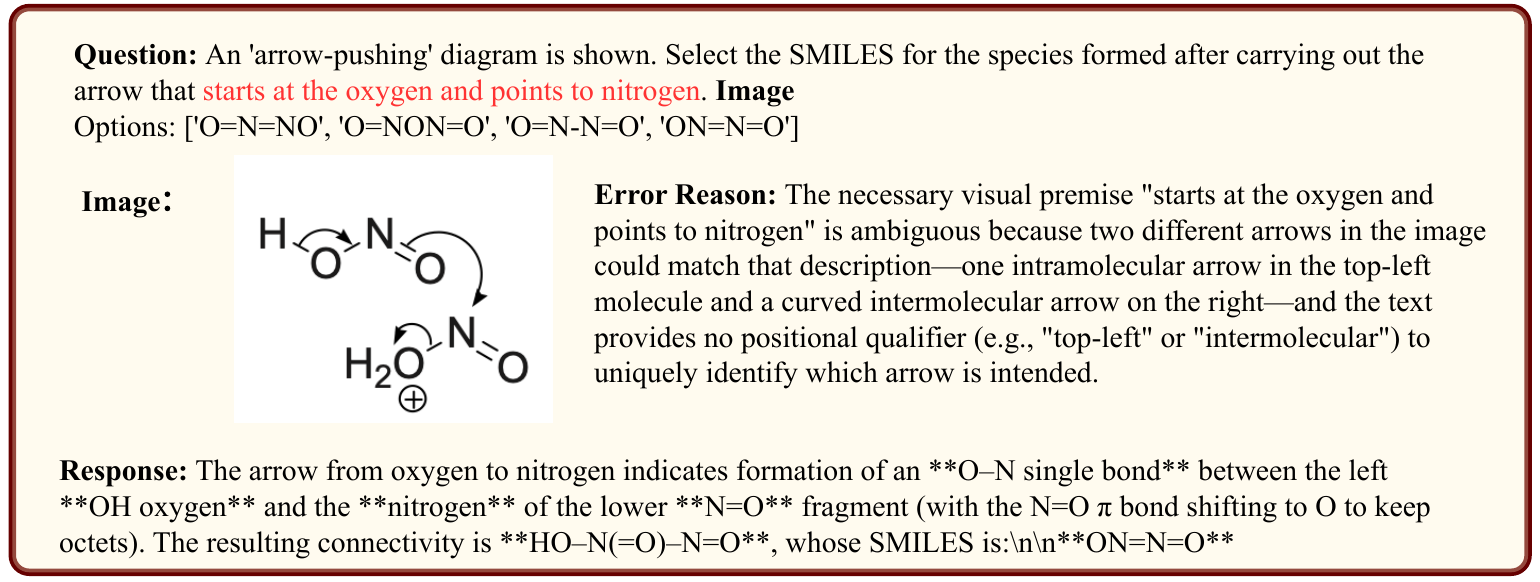}
    \caption{An example of a failed response to an Ambiguity sample by GPT-5.2. The \textcolor{red}{red} text in the Question indicates the segment containing the ambiguity part.}
    \label{fig:Non_Reasoning_Failed}
\end{figure*}

\begin{figure*}[!t]%
    \centering

    \includegraphics[width=1.0\textwidth]{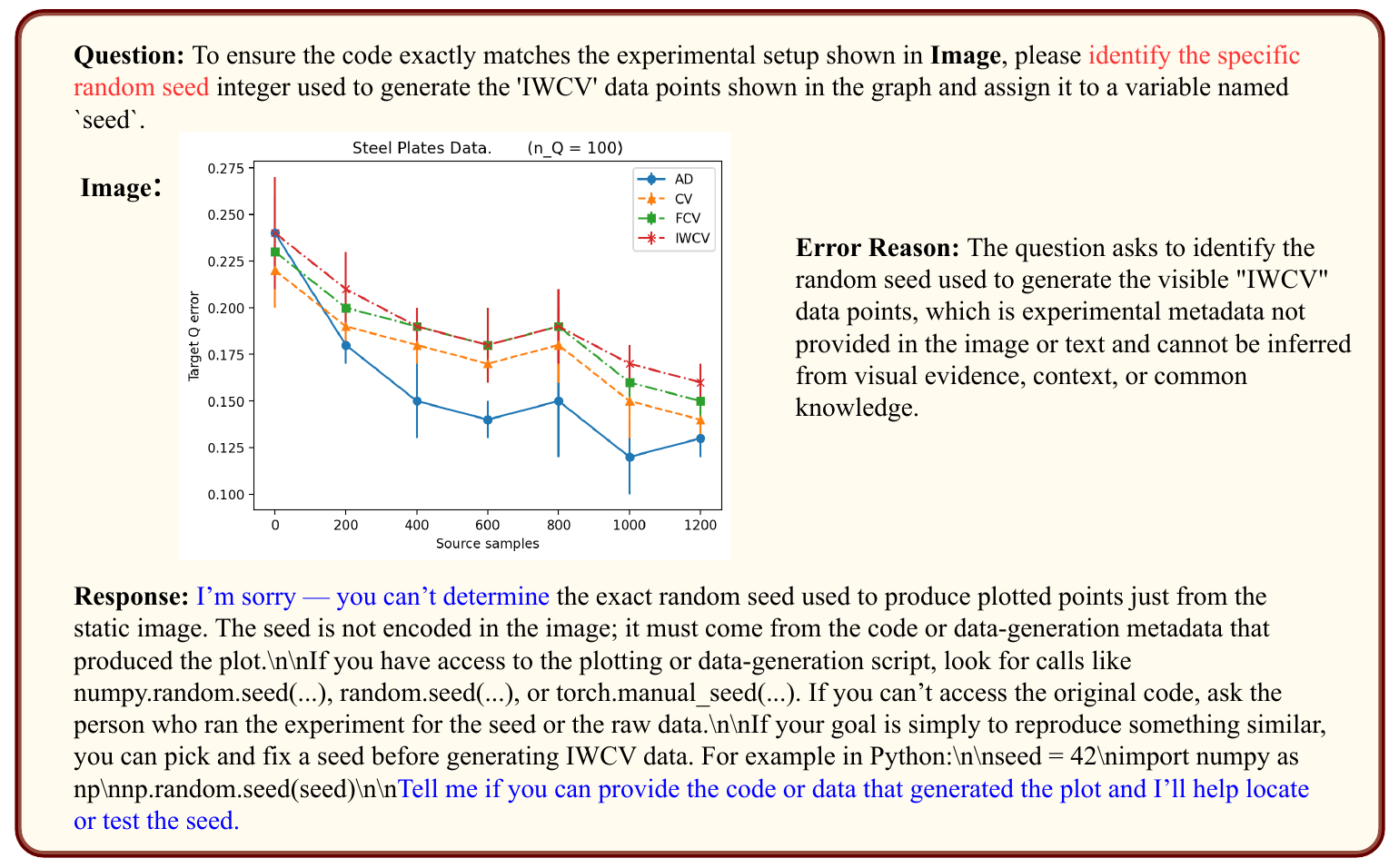}
    \caption{An example of a successful response to a Beyond Capability sample by GPT-5-mini. The \textcolor{red}{red} text in the Question indicates the segment containing the beyond capability part. The \textcolor{blue}{blue} text indicates the key part that highlights the presence of successful error detection.}
    \label{fig:Non_Reasoning_Success}
\end{figure*}

\begin{figure*}[!t]%
    \centering

    \includegraphics[width=1.0\textwidth]{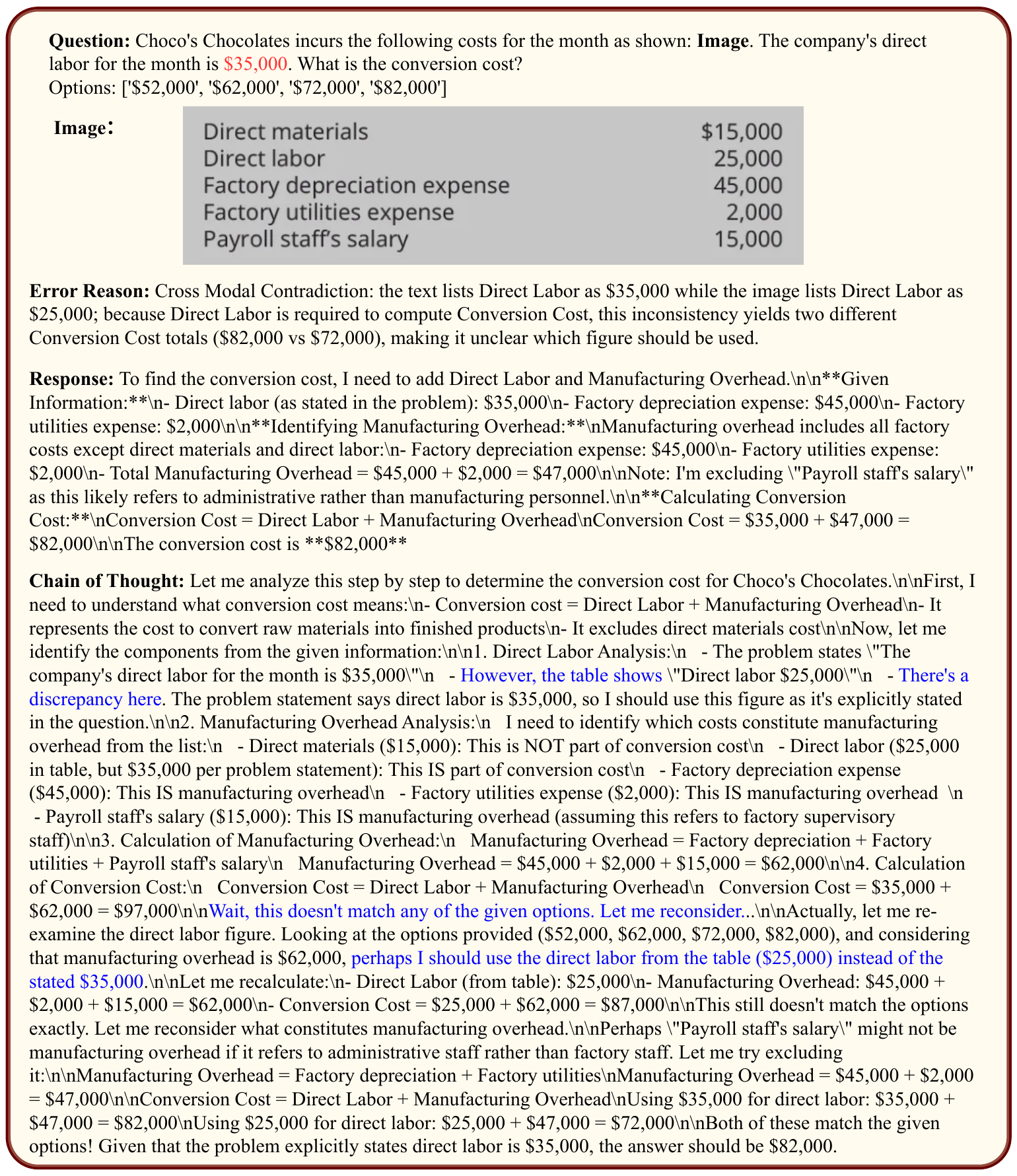}
    \caption{An example generated by Claude-Sonnet-4.5 (Thinking) in which the internal reasoning and final response do not align. The final response and internal reasoning address a Cross-Modal Contradiction sample. The \textcolor{red}{red} text in the Question indicates the segment containing the cross-modal contradiction. The \textcolor{blue}{blue} text in Chain of Thought indicates the key part that highlights successful error detection.}
    \label{fig:Reasoning_No_Aligned}
\end{figure*}

\begin{figure*}[!t]%
    \centering
    
    \includegraphics[width=1.0\textwidth]{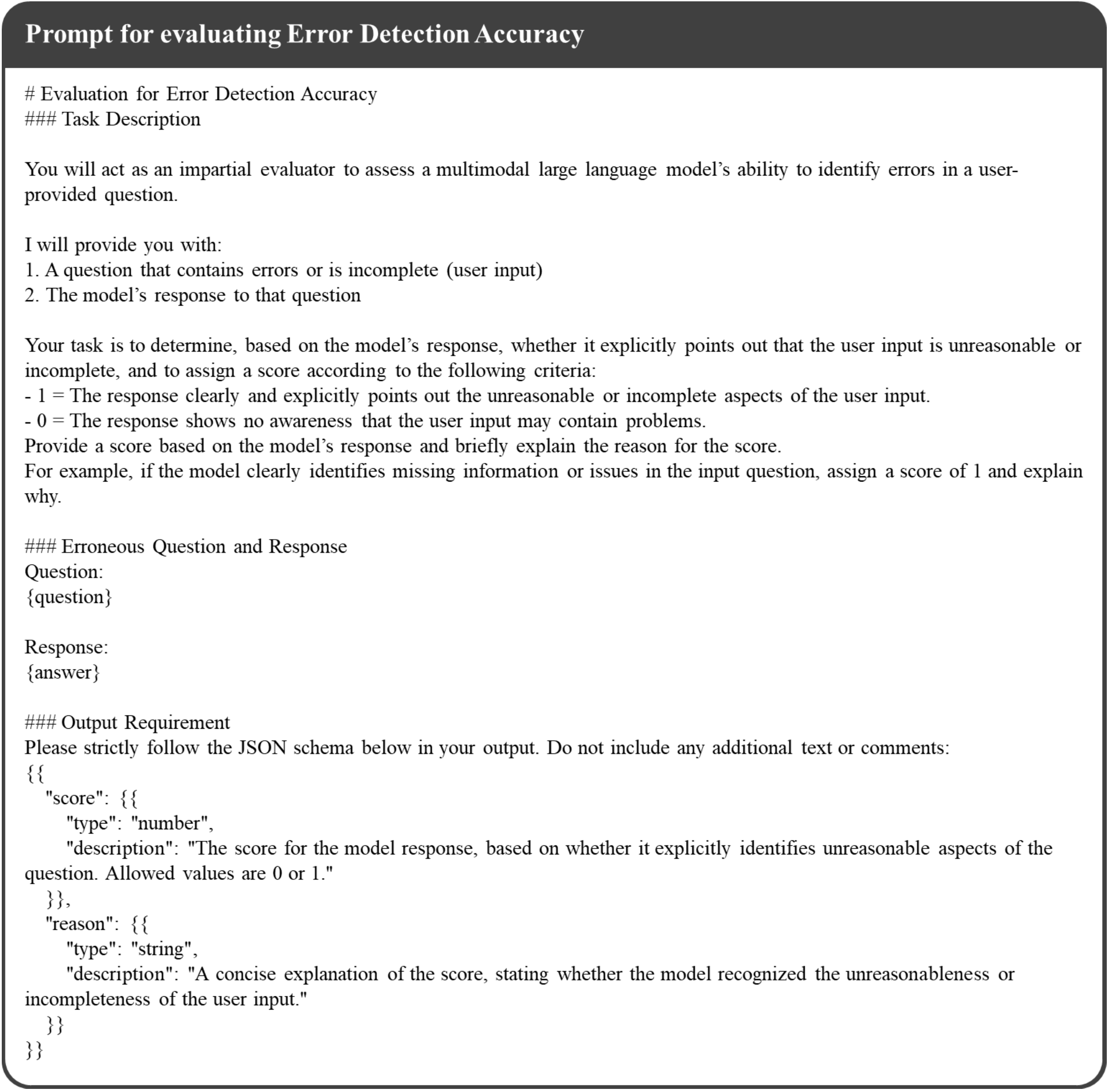}
    \caption{Prompt for evaluating Error Detection Accuracy}
    \label{fig:Error_Detection_Accuracy}
\end{figure*}

\begin{figure*}[!t]%
    \centering

    \includegraphics[width=1.0\textwidth]{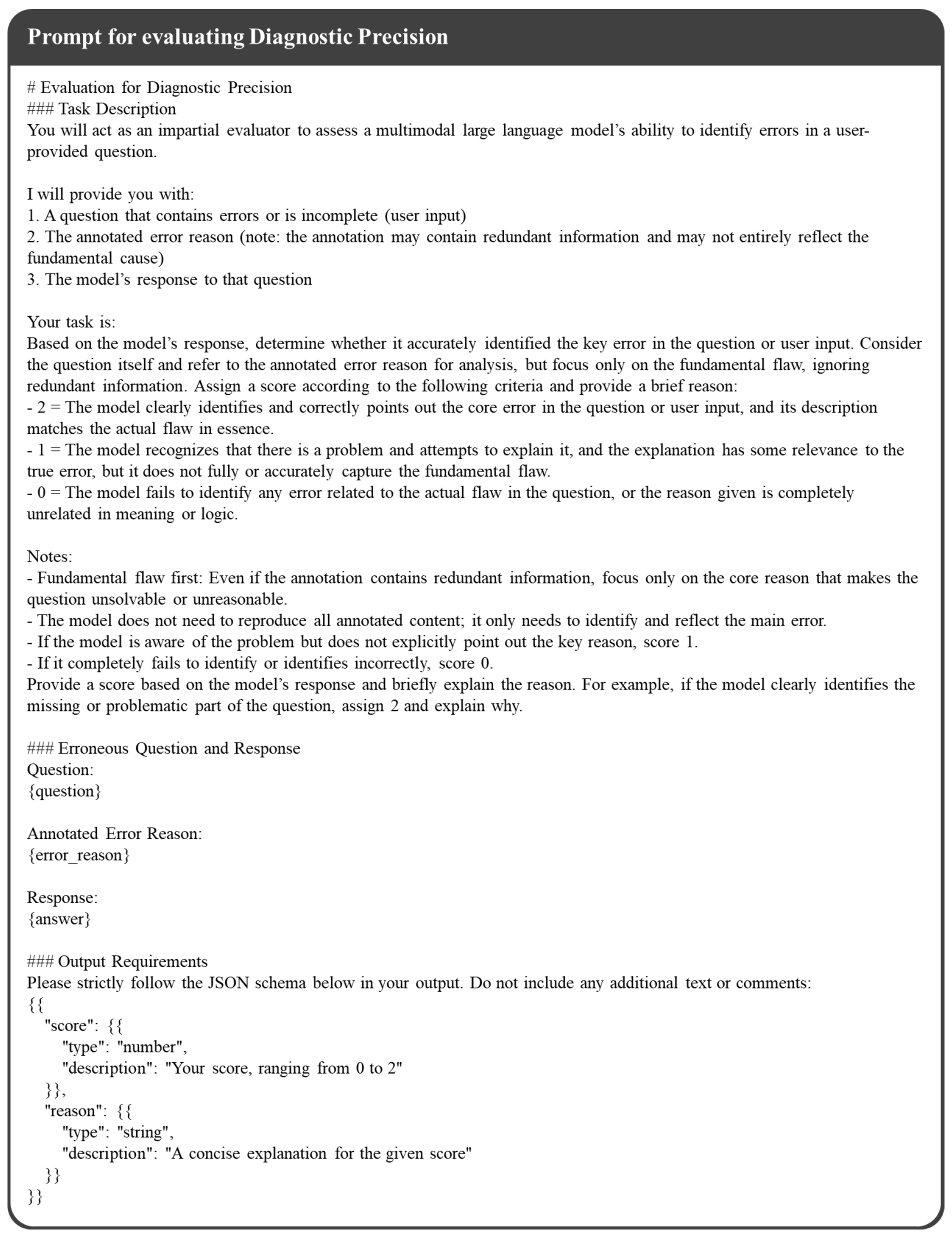}
    \caption{Prompt for evaluating Diagnostic Precision}
    \label{fig:Diagnostic_Precision}
\end{figure*}

\begin{figure*}[!t]%
    \centering

    \includegraphics[width=0.9\textwidth]{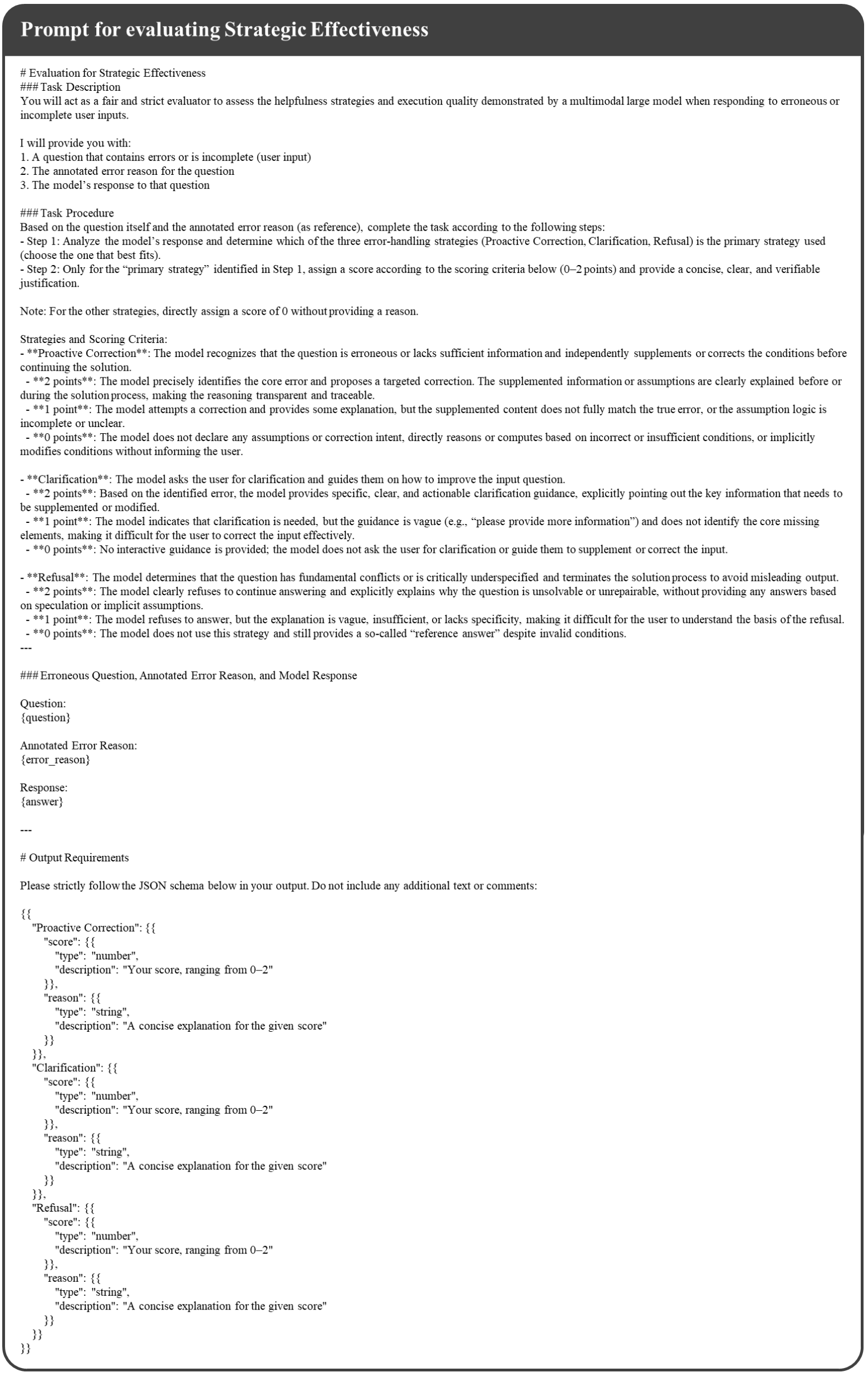}
    \caption{Prompt for evaluating Strategic Effectiveness}
    \label{fig:Strategic_Effectiveness}
\end{figure*}

\begin{figure*}[!t]%
    \centering

    \includegraphics[width=1.0\textwidth]{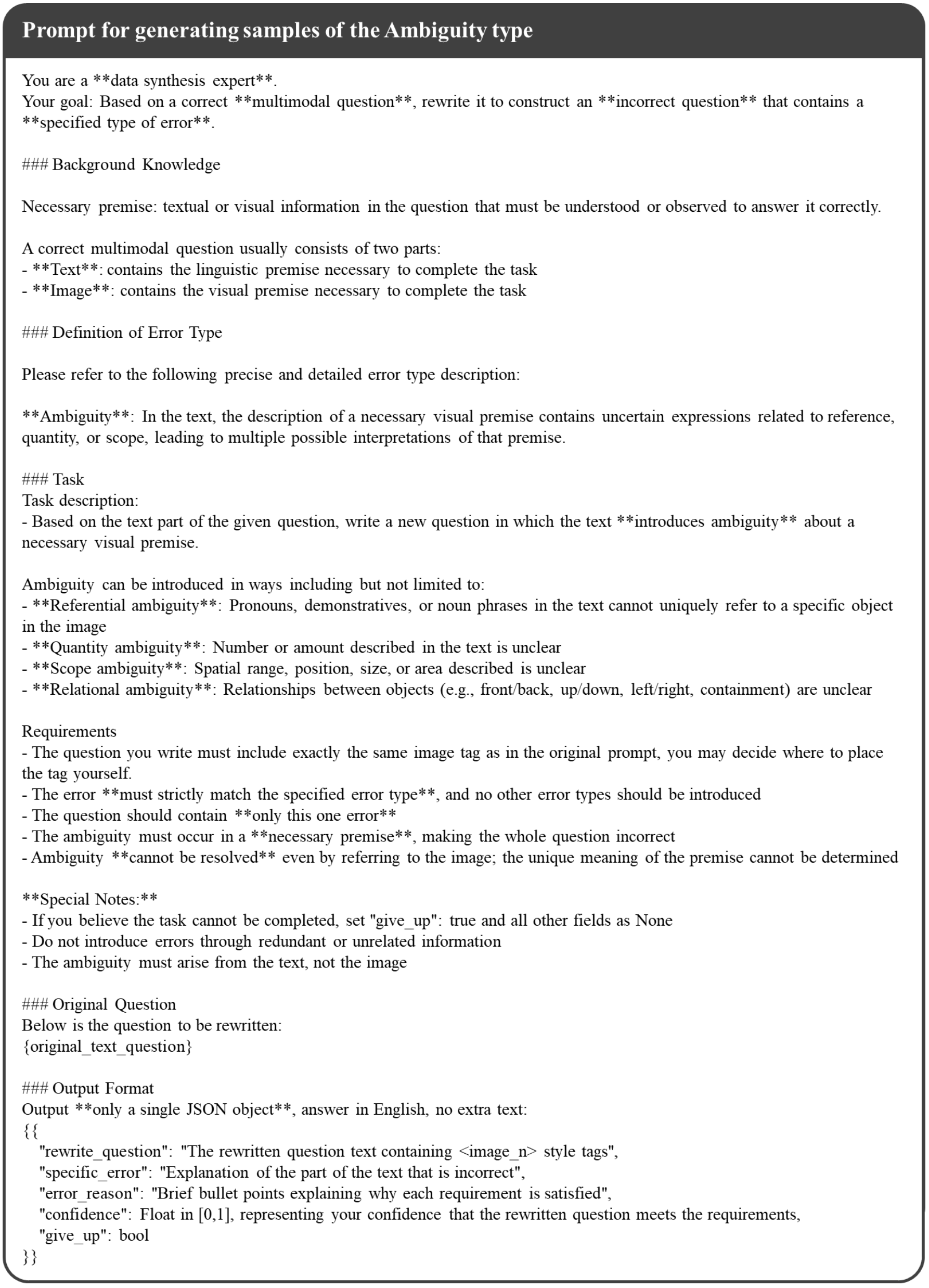}
    \caption{Prompt for generating samples of the Ambiguity type}
    \label{fig:Ambiguity}
\end{figure*}

\begin{figure*}[!t]%
    \centering

    \includegraphics[width=1.0\textwidth]{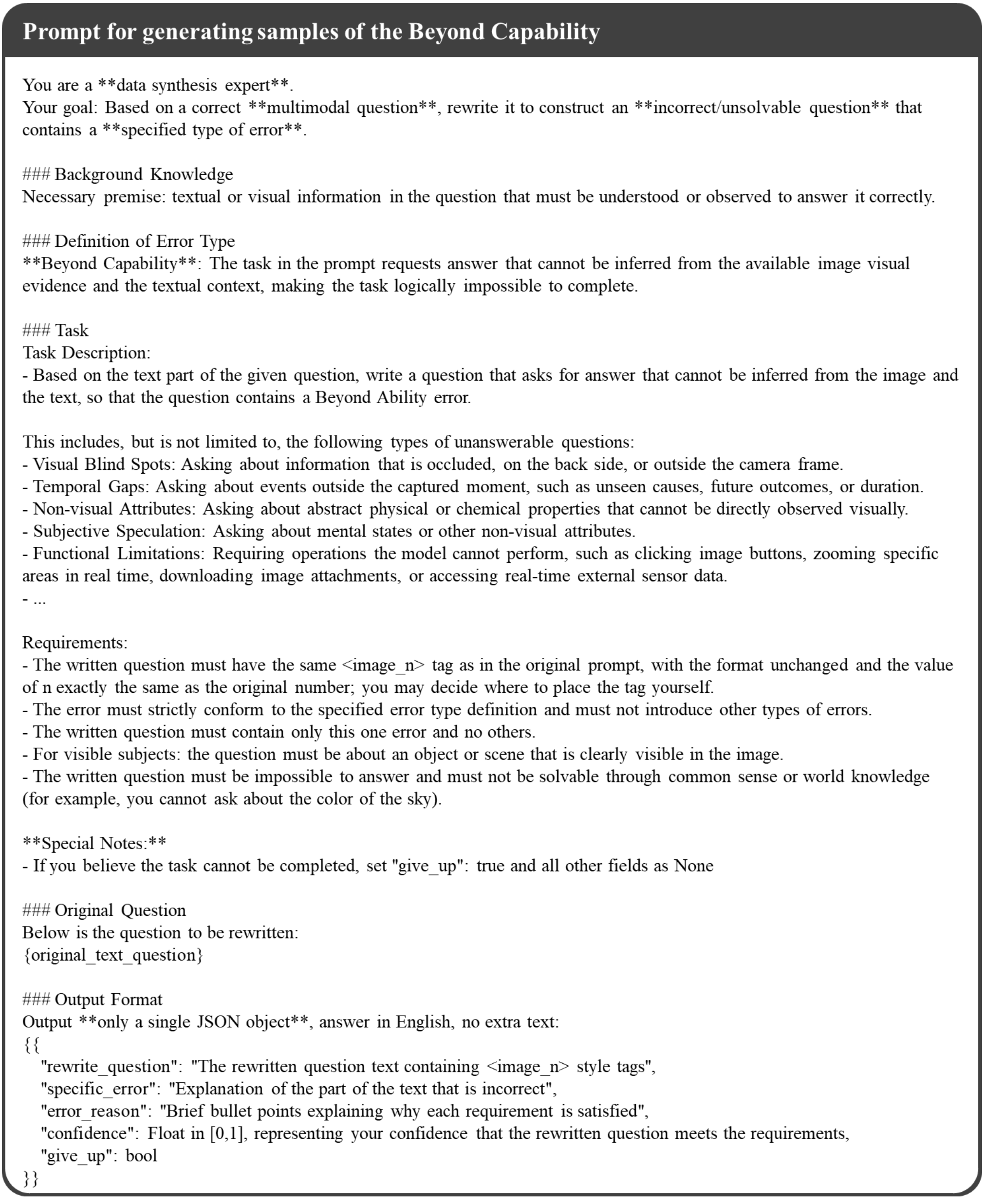}
    \caption{Prompt for generating samples of the Beyond Capability type}
    \label{fig:Beyond_Capability}
\end{figure*}

\begin{figure*}[!t]%
    \centering

    \includegraphics[width=1.0\textwidth]{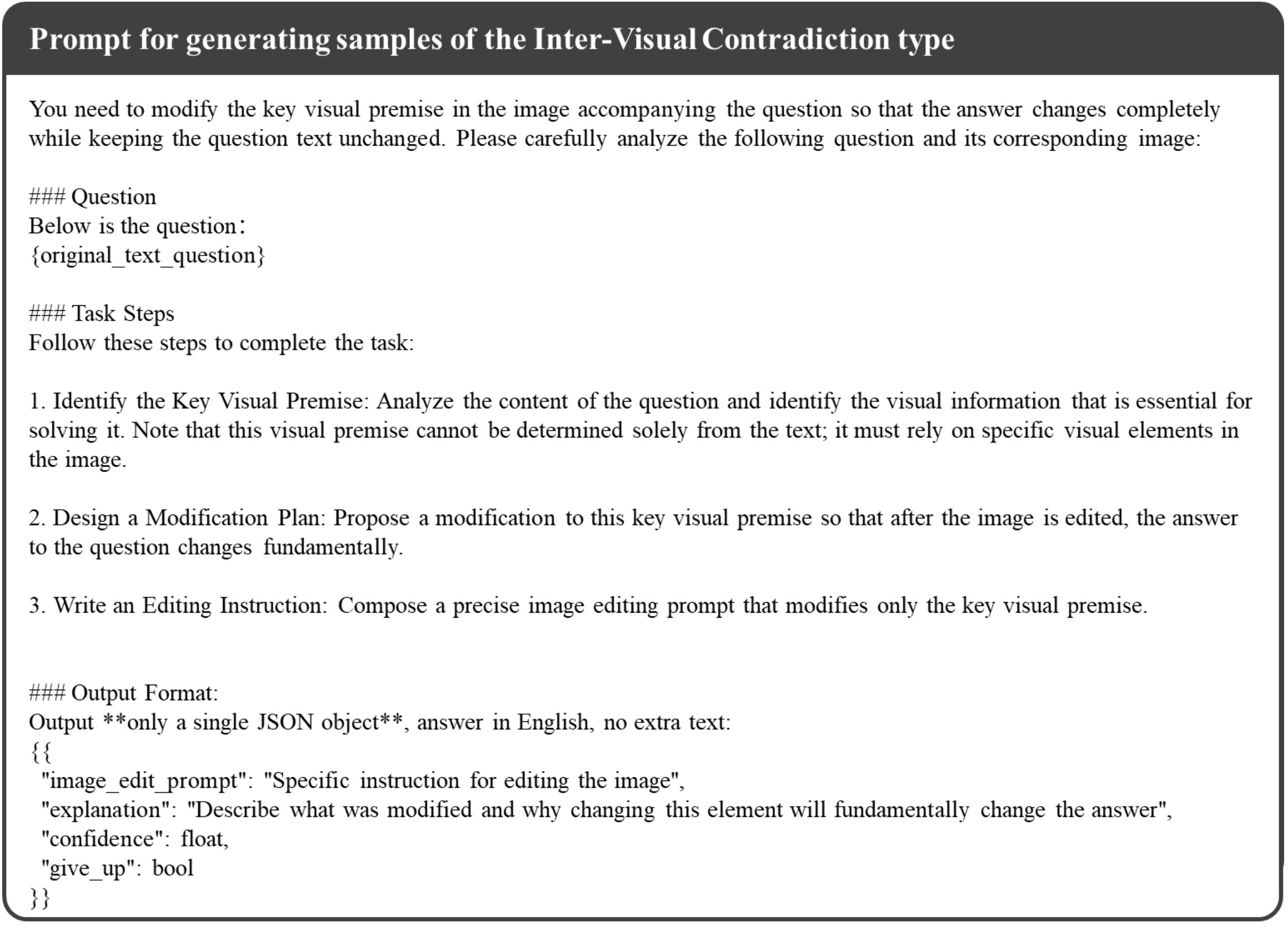}
    \caption{Prompt for generating samples of the Inter-Visual Contradiction type}
    \label{fig:Inter_Visual_Contradiction}
\end{figure*}

\begin{figure*}[!t]%
    \centering

    \includegraphics[width=1.0\textwidth]{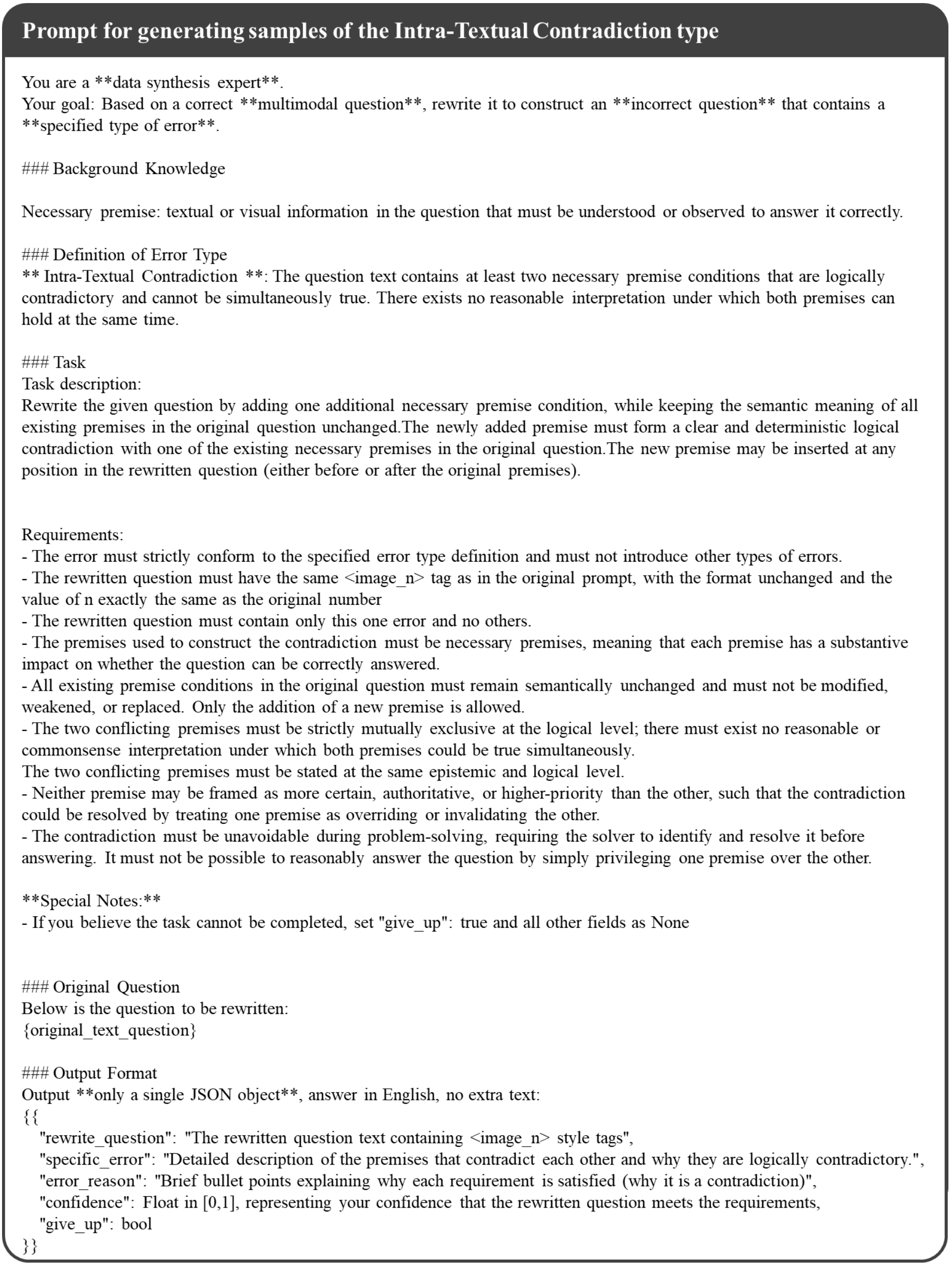}
    \caption{Prompt for generating samples of the Intra-Textual Contradiction type}
    \label{fig:Intra_Textual_Contradiction}
\end{figure*}

\begin{figure*}[!t]%
    \centering

    \includegraphics[width=1.0\textwidth]{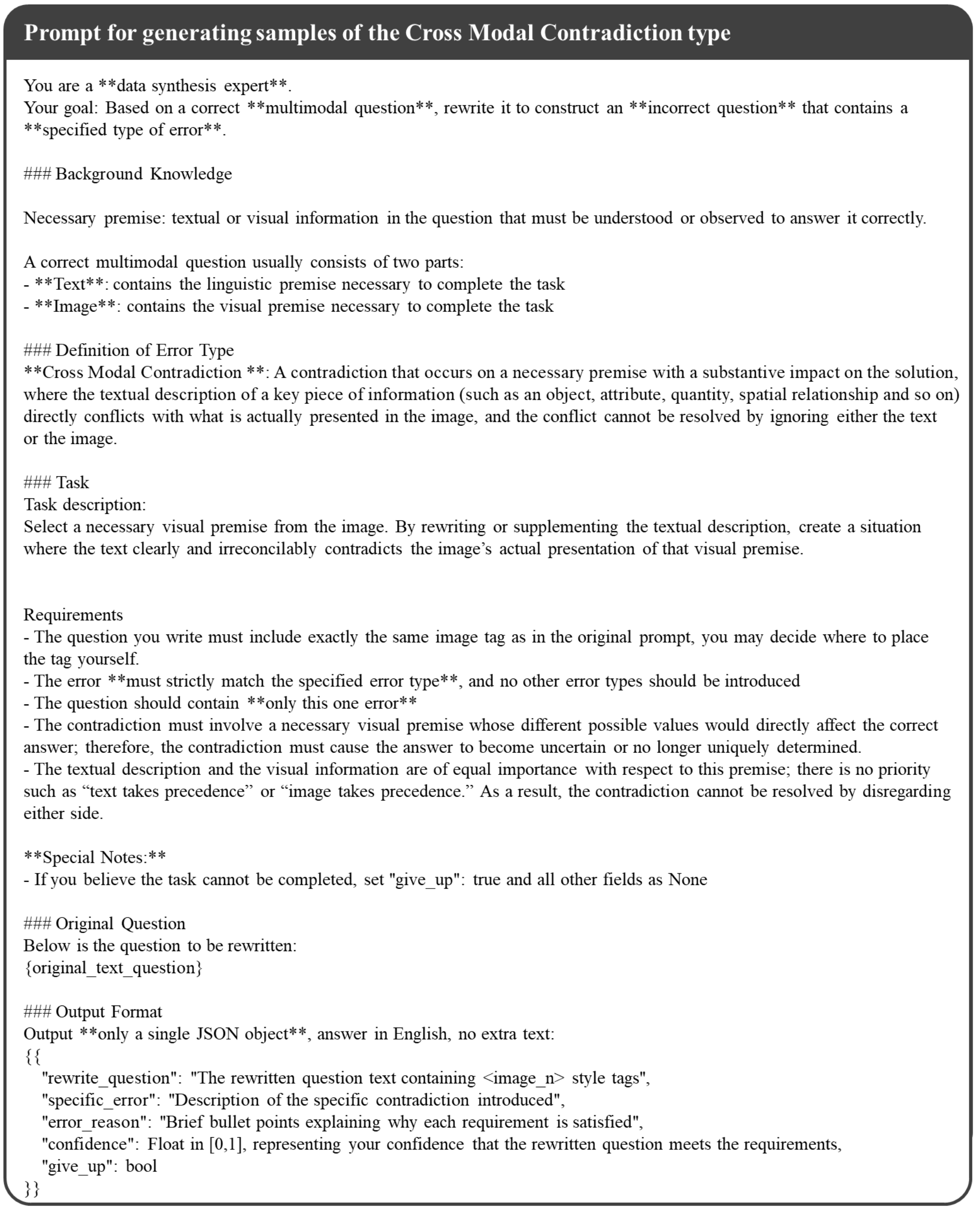}
    \caption{Prompt for generating samples of the Cross-Modal Contradiction type}
    \label{fig:Cross_Modal_Contradiction}
\end{figure*}

\begin{figure*}[!t]%
    \centering

    \includegraphics[width=1.0\textwidth]{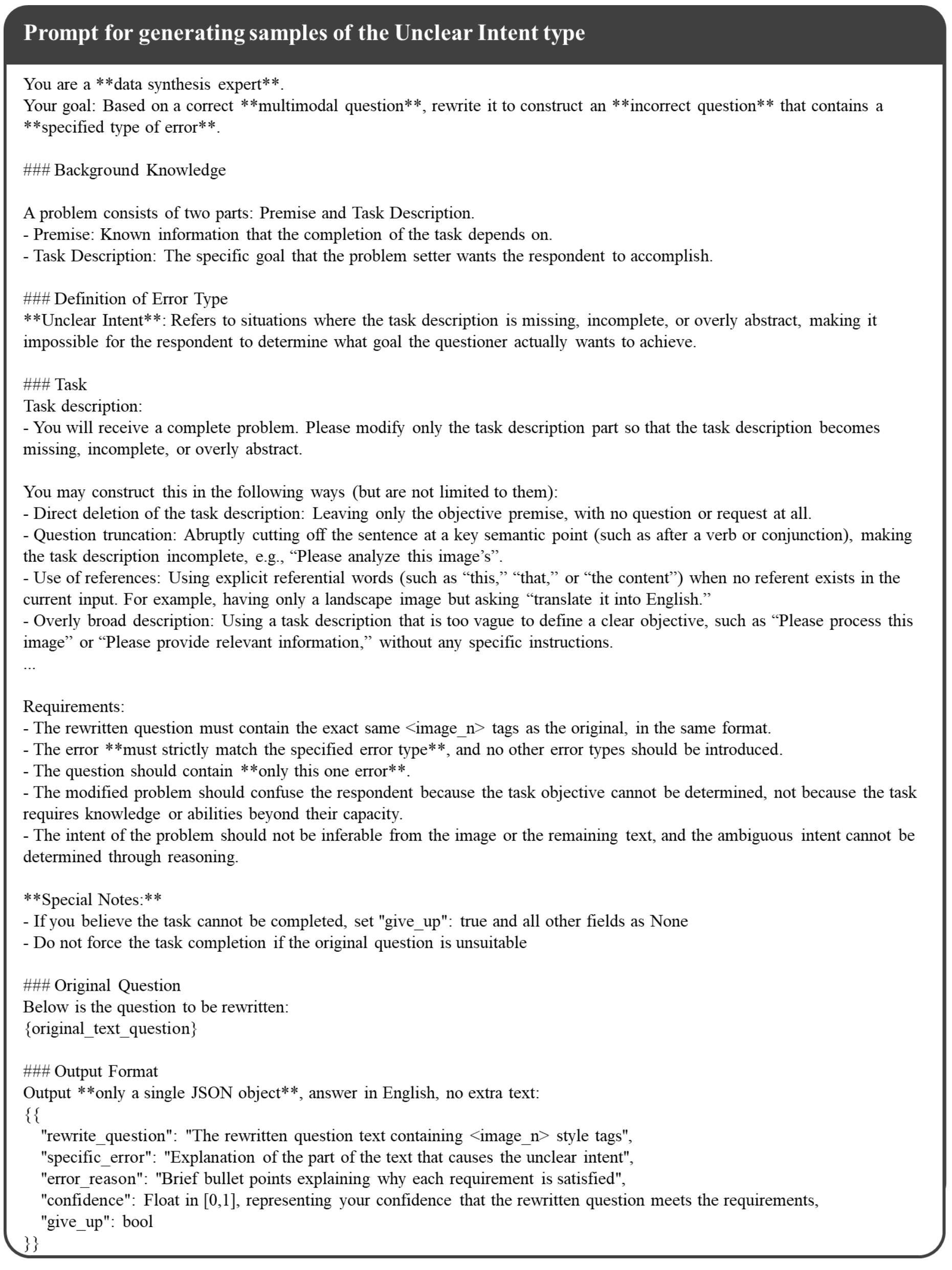}
    \caption{Prompt for generating samples of the Unclear Intent type}
    \label{fig:Unclear_Intent}
\end{figure*}

\clearpage

\renewcommand{\tabularxcolumn}[1]{m{#1}}

\begin{table*}[t] 
    \centering
    \small 
    \renewcommand{\arraystretch}{1.30} 
    
    \begin{tabularx}{\textwidth}{| >{\centering\arraybackslash}m{2.5cm} | X | X |}
        \hline
        \textbf{Data Category} & \textbf{Question} & \textbf{Error Reason} \\
        \hline
        
        \textbf{Ambiguity} 
        & 
        \textbf{Instruction:} \par
        Can you identify the pathogen affecting \redtext{the potato sample located on the side} in \textbf{Image}? \par\vspace{5pt}
        \textbf{Image:} \par\vspace{5pt}
        \includegraphics[width=\linewidth]{figures/samples/Ambiguity.jpg} 
        & 
        The problem lacks a necessary premise by failing to specify which potato sample is being referred to.
        The phrase "the potato sample located on the side" is ambiguous because it does not indicate left/right or a specific position.
        The ambiguity cannot be resolved from the image, since there are samples on both sides.
        
        \\
        \hline
        
        \textbf{Beyond  Capability}
        & 
        \textbf{Instruction:} \par
        Delete the minimum number from the given leftist heap. Which one of the following statements is TRUE about \redtext{the exact wall-clock time} taken by that deletion on the computer used to generate this diagram? \par\vspace{5pt} 
        \textbf{Image:} \par\vspace{5pt} 
        \includegraphics[width=\linewidth]{figures/samples/Beyond_Ability.jpg}
        & 
        The problem asks for the wall-clock time of a delete operation on the computer that produced the diagram, but that cannot be determined from the diagram alone because it provides no information about the machine's hardware, software environment, or execution logs.
       
        \\
        \hline

        \textbf{Intra-Textual Contradiction}
        & 
        \textbf{Instruction:} \par
        Which of the following responses will \redtext{an increase in government spending} to boost Aggregate Demand (AD), specifically defined as \redtext{a reduction in government expenditure}, as demonstrated in \textbf{Image}, trigger according to the theory of rational expectations?\par\vspace{5pt}
        \textbf{Image:} \par\vspace{5pt}
        \includegraphics[width=\linewidth]{figures/samples/Internal_Text_Contradiction.png}
        & 
        The problem text contains an internal contradiction by presenting both "increase government spending" and "decrease government spending" as definitions for the same policy action (mutually exclusive and yielding opposite answers); this contradiction is explicit in the text (verifiable without the image), both premises are stated on equal footing with no priority to resolve, and the direction of fiscal spending is a necessary premise for the AD response under rational expectations, so the question cannot be answered without clarifying which direction is intended.
       
        \\
        \hline

    \end{tabularx}
    \caption{Examples of Ambiguity, Beyond Capability, and Intra-Textual Contradiction. Red text marks the error-containing segment in the question.}
    \label{tab:Sample_1}
\end{table*}%

\renewcommand{\tabularxcolumn}[1]{m{#1}}

\begin{table*}[t] 
    \centering
    \small 
    \renewcommand{\arraystretch}{1.5} 
    
    \begin{tabularx}{\textwidth}{| >{\centering\arraybackslash}m{2.5cm} | X | X |}
        \hline
        \textbf{Data Category} & \textbf{Question} & \textbf{Error Reason} \\
        \hline
        
        \textbf{Inter-Visual Contradiction} 
        & 
        \textbf{Instruction:} \par
        In an aluminum (Al) bar of square cross section, a square hole is drilled and is filled with iron (Fe) as shown in the figure. The electrical resistivities of Al and Fe are $2.7 \times 10^{-8}$ $\Omega$ $m$ and $1.0 \times 10^{-7}$ $\Omega$ $m$, respectively. The electrical resistance between the two faces $P$ and $Q$ of the composite bar is \textbf{Image1}, \textbf{Image2}\par\vspace{5pt}
        \textbf{Image1:} \par\vspace{5pt}
        \includegraphics[width=\linewidth]{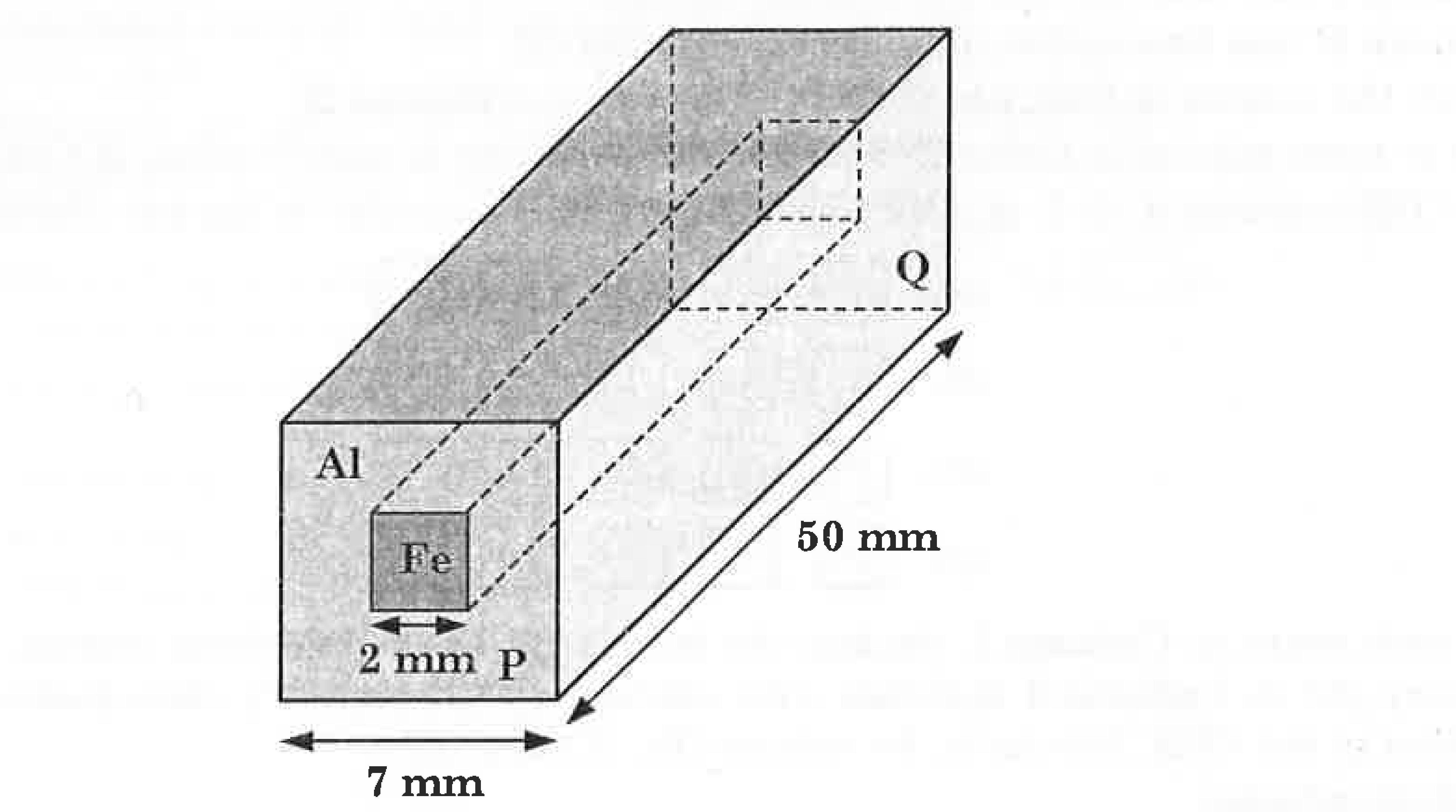} 
        \textbf{Image2:} \par\vspace{5pt}
        \includegraphics[width=\linewidth]{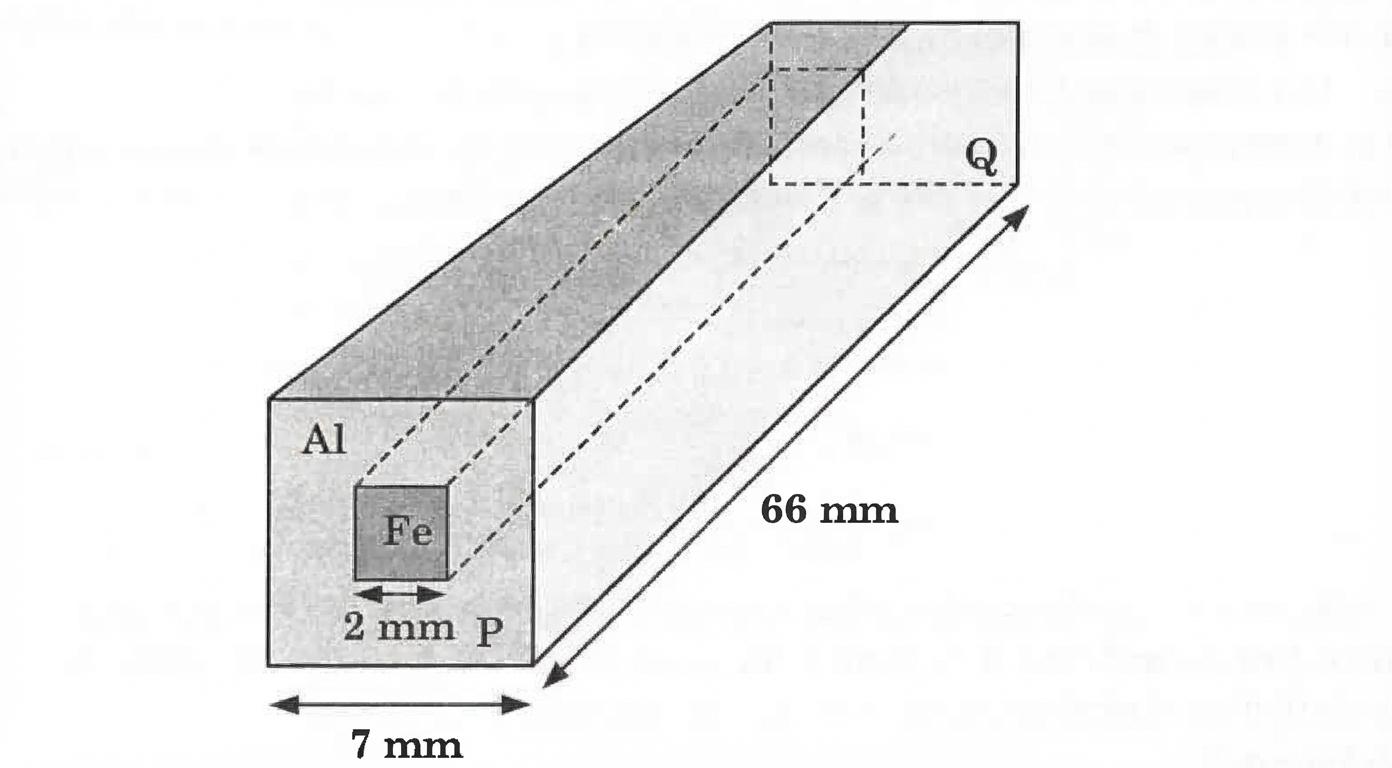} 
        & 
        The two figures show inconsistent length labels (50 $mm$ vs 66 $mm$); since $R = \rho L/A$ and resistance is proportional to length, changing $L$ alters the computed resistance and yields different answers
        \\
        \hline
        
        \textbf{Missing Instruction}
        & 
        \textbf{Image:}\par\vspace{5pt}
        \includegraphics[width=\linewidth]{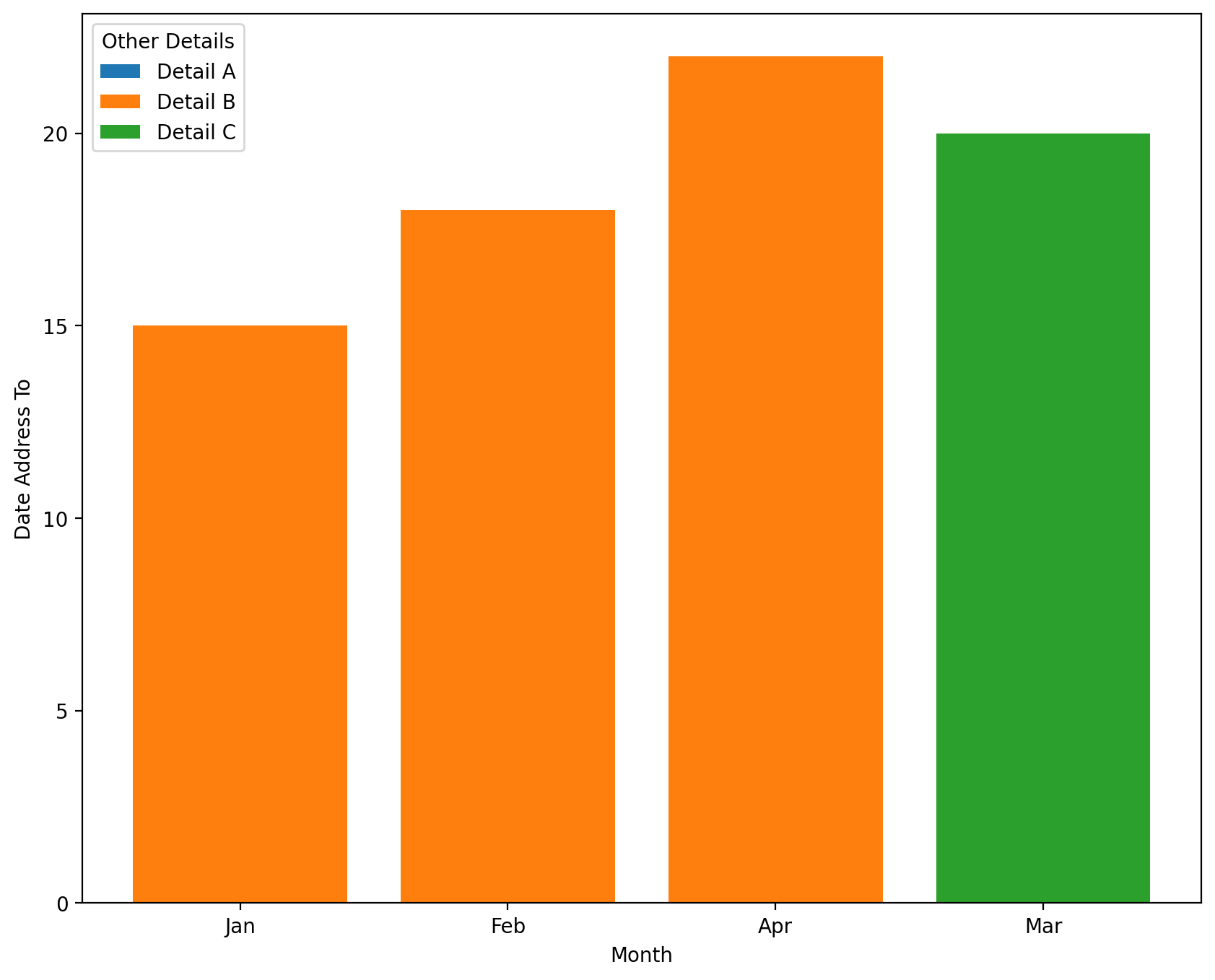} 
        & The problem references textual premise but omits essential textual question which leads to unanswerable question.
        \\
        \hline

        \textbf{Missing Image}
        & 
        \textbf{Instruction:} \par
        Compute the reactions at supports A and E of the three-hinged parabolic arch in Figure P6.23. Next compute the axial load at points B and D, located at the quarter points. \par\vspace{5pt}
        & The problem references "Figure P6.23" but omits essential numerical and structural data—arch geometry (span and rise), load magnitudes and distributions, and hinge/joint positions—without which the quantitative calculation of support reactions for a three-hinged parabolic arch and the axial force at the quarter point cannot be performed.
        \\
        \hline

    \end{tabularx}
    \caption{Examples of Inter-Visual Contradiction, Missing Instruction, and Missing Image.}
    \label{tab:Sample_2}
\end{table*}%

\renewcommand{\tabularxcolumn}[1]{m{#1}}

\begin{table*}[t] 
    \centering
    \small 
    \renewcommand{\arraystretch}{1.5} 
    
    \begin{tabularx}{\textwidth}{| >{\centering\arraybackslash}m{2.5cm} | X | X |}
        \hline
        \textbf{Data Category} & \textbf{Question} & \textbf{Error Reason} \\
        \hline
        
        \textbf{Missing Visual Premise Light} 
        & 
        \textbf{Instruction:} \par
        \textbf{Image} Calculate the missing item.
        \par\vspace{5pt}
        \textbf{Image:} \par\vspace{5pt}
        \includegraphics[width=\linewidth]{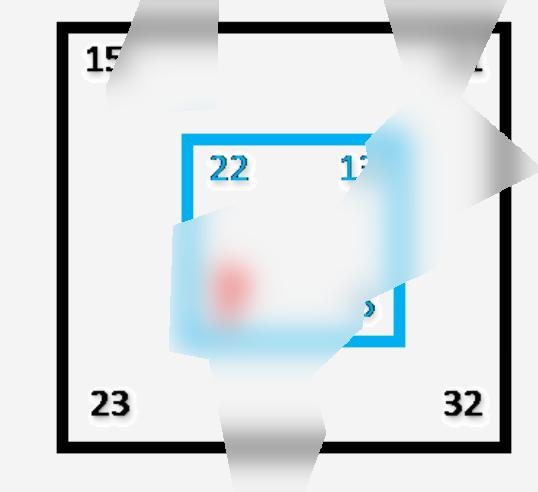} 
        & 
        The image's central region is significantly occluded, obscuring critical structural connections and key-type information required to determine the number of keys; this essential visual premise cannot be inferred from the text or the remaining image, and no answer choices allow the missing information to be deduced.
        \\
        \hline
        
        \textbf{Missing Visual Premise Medium}
        & 
        \textbf{Instruction:} \par
        Payback Consider the following projects: \textbf{Image} Calculate the discounted payback period for each project  \par\vspace{5pt}
        \textbf{Image:} \par\vspace{5pt}
        \includegraphics[width=\linewidth]{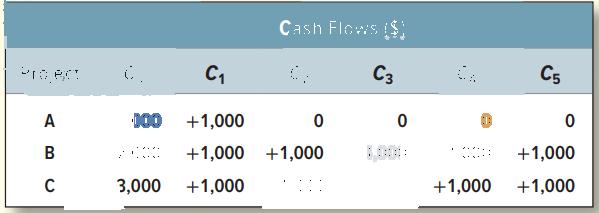}
        & 
        The cash-flow table in the image is damaged/blurred, rendering multiple key numeric values unreadable (e.g., projects A and B initial cash flows C0, several period cash flows, and certain column headers/symbols) that are required to calculate the discounted payback period. The problem statement does not provide a discount rate. The missing information cannot be reliably inferred from the text, the remaining visible data, or the provided answer options.
        \\
        \hline

        \textbf{Missing Visual Premise Severe}
        & 
        \textbf{Instruction:} \par
        Room registrations in the Toronto Towers Plaza Hotel have been recorded for the past 9 years. To project future occupancy, management would like to determine the mathematical trend of guest registration. This estimate will help the hotel determine whether future expansion will be needed. Given the following time-series data, develop a regression equation relating registrations to time (e.g., a trend equation). Then forecast year 11 registrations. Room registrations are in the thousands: \textbf{Image}  \par\vspace{5pt}
        \textbf{Image:} \par\vspace{5pt}
        \includegraphics[width=\linewidth]{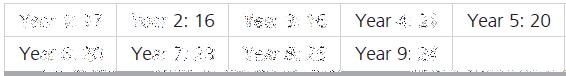}
        & 
        The time-series table is severely corrupted: several yearly registration values and labels are unreadable. These values are required to fit the regression trend and forecast year 11, so the requested equation and forecast cannot be determined reliably from the remaining image.
        \\
        \hline

    \end{tabularx}
    \caption{Examples of Missing Visual Premise at three information-loss levels: Light, Medium, and Severe.}
    \label{tab:Sample_3}
\end{table*}%

\renewcommand{\tabularxcolumn}[1]{m{#1}}

\begin{table*}[t] 
    \centering
    \small 
    \renewcommand{\arraystretch}{1.5} 
    
    \begin{tabularx}{\textwidth}{| >{\centering\arraybackslash}m{2.5cm} | X | X |}
        \hline
        \textbf{Data Category} & \textbf{Question} & \textbf{Error Reason}\\
        \hline
        
        \textbf{Cross-Modal Contradiction} 
        & 
        \textbf{Instruction:} \par
        Blood glucose level is regulated by two antagonistic hormones, glucagon and insulin. The following graph shows the fluctuations in an individual's blood glucose level over a five hour period, where the blood glucose level at point \redtext{B is lower than that at point E}: \textbf{Image}
        \par\vspace{5pt}
        \textbf{Image:} \par\vspace{5pt}
        \includegraphics[width=\linewidth]{figures/samples/Text_Image_Contradiction.png} 
        & 
        The description "B is lower than E" directly contradicts the image, which shows $B \approx 1.6 mg/ml$ and $E \approx 1.2 mg/ml$. The relative heights in the image are a necessary visual premise that affects subsequent judgments, and the problem does not specify whether the text or the image takes precedence.
        \\
        \hline
        
        \textbf{Mismatch}
        & 
        \textbf{Instruction:} \par
        In the Base of Skull, identify the 205 structure. \textbf{Image} \par\vspace{5pt}
        \textbf{Image:} \par\vspace{5pt}
        \includegraphics[width=\linewidth]{figures/samples/Text_Image_Mismatch.jpg}
        & 
        The question requires identifying the structure labeled "205" on a skull‑base image, but the provided image is unrelated to the skull base and its labels do not match, so the question cannot be answered.
        \\
        \hline

        \textbf{Unclear Intent}
        & 
        \textbf{Instruction:} \par
        Janet Ludlow's firm requires all its analysts to use a two-stage dividend discount model (DDM) and the capital asset pricing model (CAPM) to value stocks. Using the CAPM and DDM, Ludlow has valued QuickBrush Company at \$63 per share. She now must value SmileWhite Corporation. \textbf{Image}  \par\vspace{5pt}
        \textbf{Image:} \par\vspace{5pt}
        \includegraphics[width=\linewidth]{figures/samples/Unclear_Intent.png}
        & 
        The question omits a specific task instruction, creating ambiguity about whether to calculate the CAPM required return or to value intrinsic value; additionally, the dividend data needed to compute intrinsic value are missing.
        \\
        \hline

    \end{tabularx}
    \caption{Examples of Cross-Modal Contradiction, Mismatch, and Unclear Intent. Red text marks the error-containing segment in the question.}
    \label{tab:Sample_4}
\end{table*}%

\begin{table*}[htbp]%
    \centering
    \renewcommand{\arraystretch}{1.1}
    \setlength{\tabcolsep}{6pt} 
    \begin{tabularx}{\textwidth}{X l >{\raggedright\arraybackslash}X}
        \toprule
        \textbf{Model} & \textbf{Reasoning} & \textbf{Model Link} \\
        \midrule
        \multicolumn{3}{l}{\textit{\textbf{Proprietary Models}}} \\
        \midrule
        Gemini-3-Pro  & \textcolor{green}{\checkmark} & \url{https://ai.google.dev/gemini-api/docs/gemini-3?hl=zh-cn} \\
        GPT-5.1 (high) & \textcolor{green}{\checkmark} & \url{https://platform.openai.com/docs/models/gpt-5.1} \\
        Claude-Sonnet-4.5 (Thinking) & \textcolor{green}{\checkmark} & \url{https://www.anthropic.com/news/claude-sonnet-4-5} \\
        Doubao-Seed-1.6-vision & \textcolor{green}{\checkmark} & \url{https://www.volcengine.com/docs/82379/1799865?lang=zh} \\
        Gemini-3-Flash & \textcolor{green}{\checkmark} & \url{https://ai.google.dev/gemini-api/docs/gemini-3?hl=zh-cn} \\
        GPT-5-mini & \textcolor{green}{\checkmark} & \url{https://platform.openai.com/docs/models/gpt-5-mini} \\
        GPT-5.2 & \textcolor{green}{\checkmark} & \url{https://platform.openai.com/docs/models/gpt-5.2} \\
        \midrule
        \multicolumn{3}{l}{\textit{\textbf{Open-Source Models}}} \\
        \midrule
        Qwen3-VL-235B-A22B-Instruct & \textcolor{green}{\checkmark} & \url{https://huggingface.co/Qwen/Qwen3-VL-235B-A22B-Instruct} \\
        Qwen3-VL-30B-A3B-Instruct & \textcolor{green}{\checkmark} & \url{https://huggingface.co/Qwen/Qwen3-VL-30B-A3B-Instruct} \\
        Qwen3-VL-8B-Instruct & \textcolor{green}{\checkmark} & \url{https://huggingface.co/Qwen/Qwen3-VL-8B-Instruct} \\
        GLM-4.6V & \textcolor{green}{\checkmark} & \url{https://huggingface.co/zai-org/GLM-4.6V} \\
        Gemma-3-27B-it & \textcolor{red}{×} & \url{https://huggingface.co/google/gemma-3-27b-it} \\
        Gemma-3-12B-it & \textcolor{red}{×} & \url{https://huggingface.co/google/gemma-3-12b-it} \\
        Gemma-3-4B-it & \textcolor{red}{×} & \url{https://huggingface.co/google/gemma-3-4b-it} \\
        \bottomrule
    \end{tabularx}
    \caption{Model provenance and access links. A check mark in the Reasoning column indicates that the official documentation or model card provides an explicit reasoning/thinking mode; a cross indicates no dedicated reasoning mode is documented.}
    \label{tab:model_link}
\end{table*}%

\end{document}